\DocumentMetadata{%
	pdfstandard=A-3u,
	pdfversion=1.7,
	lang=en-US,
}

\documentclass[nofoot, balance,colorlinks,upint,subscriptcorrection,varvw,mathalfa=cal=boondoxo,spanish,german,vietnamese,russian,greek]{asmeconf}

\hypersetup{%
	pdfauthor={John H. Lienhard},									  
	pdftitle={ASME Conference Paper LaTeX Template},                  
	pdfkeywords={ASME conference paper, LaTeX template, BibTeX style},
	pdfsubject = {Describes the asmeconf LaTeX template},			  
	pdflicenseurl={https://ctan.org/pkg/asmeconf},
}

\usepackage{amsmath} 
\usepackage{todonotes}
\usepackage{multirow}
\usepackage{enumitem}

\usepackage{wrapfig} 
\usepackage{placeins}

\begin{document}



\SetAuthors{%
    Kristen M. Edwards\affil{1}\CorrespondingAuthor{kme@mit.edu}, 
	Farnaz Tehranchi\affil{2}, 
	Scarlett R. Miller\affil{2}, 
	Faez Ahmed\affil{1}
	}

\SetAffiliation{1}{Massachusetts Institute of Technology, Cambridge, MA}
\SetAffiliation{2}{The Pennsylvania State University, University Park, PA }

\title{AI Judges in Design: Statistical Perspectives on \\ Achieving Human Expert Equivalence With Vision-Language Models}

\maketitle 



\begin{abstract}
The subjective evaluation of early stage engineering designs, such as conceptual sketches, traditionally relies on human experts. However, expert evaluations are time-consuming, expensive, and sometimes inconsistent. Recent advances in vision-language models (VLMs) offer the potential to automate design assessments, but it is crucial to ensure that these AI ``judges'' perform on par with human experts. 
However, no existing framework assesses expert equivalence. This paper introduces a rigorous statistical framework to determine whether an AI judge's ratings match those of human experts. We apply this framework in a case study evaluating four VLM-based judges on key design metrics (uniqueness, creativity, usefulness, and drawing quality). These AI judges employ various in-context learning (ICL) techniques, including uni- vs. multimodal prompts and inference-time reasoning. The same statistical framework is used to assess three trained novices for expert--equivalence.
Results show that the top-performing AI judge, using text- and image-based ICL with reasoning, achieves expert-level agreement for uniqueness and drawing quality and outperforms or matches trained novices across all metrics. In 6/6 runs for both uniqueness and creativity, and 5/6 runs for both drawing quality and usefulness, its agreement with experts meets or exceeds that of the majority of trained novices. These findings suggest that reasoning-supported VLM models can achieve human-expert equivalence in design evaluation. This has implications for scaling design evaluation in education and practice, and provides a general statistical framework for validating AI judges in other domains requiring subjective content evaluation.

\end{abstract}


\date{Version \versionno, \today}


\section{Introduction}
In engineering design, assessing the quality and creativity of early concept sketches is a critical but challenging task~\cite{christiaans2002creativity,christensen2016dimensions, sternberg1999handbook}. Designers and researchers often rely on subjective evaluations by domain experts to judge the creativity, novelty, aesthetic appeal, and functional feasibility of a design concept~\cite{baer2004extension, baer2015importance, kaufman2008essentials}. This approach, while considered the gold standard, faces well-known issues: expert assessment is labor-intensive, potentially costly, and can suffer from variability due to personal biases and differing criteria. Furthermore, defining expertise within a domain is inherently complex and remains an active area of research~\cite{kaufman2011expertise}; however, leading researchers argue that expert-level judges should possess ``at least some formal training and experience in the target domain''~\cite{amabile1996creativity}. Even within panels of experts, achieving consistency is non-trivial, typically necessitating multiple judges and statistical checks for inter-rater reliability (e.g., intraclass correlation coefficients)~\cite{miller2021should}.
At the same time, crowdsourcing has been explored to increase throughput of design evaluations, but without careful control, as it may yield noisy results~\cite{10.1145/3274384, miller2021crowdsourcing}. Studies show that averaging across a large crowd of non-experts can lead to inaccurate evaluations, with some raters providing systematically biased judgments~\cite{burnap2015crowdsourcing}. These challenges motivate the search for scalable, reliable alternatives or complements to human expert judges.

Artificial intelligence (AI) offers a promising avenue to address this need. In particular, vision-language models (VLMs) – AI models that combine image and text understanding – have recently achieved remarkable capabilities in interpreting visual content and producing human-like judgments~\cite{liu2023visualinstructiontuning, OpenAI_o1_2024, OpenAI_GPT4o_2024}. For example, DriveLLaVA~\cite{s24134113} is a VLM fine-tuned to make human-level judgments in driving scenarios, while Lin et al.~\cite{lin2024evaluatingtexttovisualgenerationimagetotext} developed models capable of distinguishing AI-generated from human-created art. Their work was driven by the observation that humans can often tell the difference, but large-scale human evaluation is costly and impractical. 

In engineering design, a sufficiently advanced AI could assess a sketch’s creativity or functionality in seconds, enabling real-time feedback and rapid screening of large idea pools. However, deploying AI as a design ``judge'' demands caution: its evaluations must be as credible and valid as those of a human expert. In other words, the AI’s ratings should be \textbf{equivalent to expert ratings} for us to trust its judgments in critical design decisions. 

Recent developments make the goal of expert equivalence increasingly plausible. Modern VLMs and other multimodal methods have demonstrated an ability to understand drawings or images and relate them to semantic concepts. In engineering design, such models have shown high correlation with human assessments of creativity and other metrics~\cite{song2023aemultimodal, edwards2021design, su2023multimodal, yuan2021multimodal}. However, these methods require large amounts of training data, limiting their practical use.

Preliminary work has shown that pre-trained VLMs can assess design similarity~\cite{picard2024conceptmanufacturingevaluatingvisionlanguage}, and other studies explore LLMs in engineering tasks like solving mechanics problems~\cite{NI2024mechagent} or building trusses using FEM modules~\cite{jadhav2024largelanguagemodelagent}. These results suggest that with the right models and inputs, AI can capture the nuanced criteria used by experts.

Yet, a key challenge remains: rigorously verifying that AI evaluators truly match expert performance. Prior studies often report high correlations or small average errors~\cite{zheng2023llmasjudge, song2023aemultimodal}. While encouraging, such measures alone do not confirm that the AI's performance is indistinguishable from that of a human expert within acceptable bounds. What is needed is a rigorous statistical equivalence perspective: instead of asking whether the AI is simply correlated with or not significantly worse than humans, we ask whether any difference between the AI and human evaluations is within acceptable bounds across multidimensional criteria. Adopting statistical equivalence testing provides a more stringent and meaningful validation that the AI can serve as a surrogate for human judges with confidence and highlights areas of improvement for future AI judges.

In this paper, we address this gap by developing AI “judges” for evaluating design sketches and introducing a comprehensive framework to test for expert equivalence. We focus on a use case of rating milk frother design sketches on subjective metrics commonly used in conceptual design assessment (e.g.,  uniqueness, creativity, usefulness, and drawing quality). We test four VLM-based models using a dataset of design sketches with scores obtained by two experts and three trained novices. We then design and apply a rigorous testing procedure to determine whether the models’ ratings are statistically on par with human expert ratings, within a predefined tolerance margin. To our knowledge, this work is the first to demonstrate and validate an AI model achieving expert-level equivalence in an engineering design evaluation context.

\textbf{Contributions:} The key contributions of this work are as follows:
\begin{itemize}[nosep]

  \item We demonstrate how \emph{in-context} VLM predictions (few-shot prompting) can be used to evaluate subjective design metrics, without conventional fine-tuning or large-scale training. This approach both addresses the scarcity of labeled data in many design contexts, while also enabling scalable design evaluation, which is necessary as the quantity of generated designs grows.
  

    
\item We propose a \emph{comprehensive statistical validation framework} to assess AI-expert agreement across multiple dimensions. Our approach integrates correlation metrics, distributional-difference hypothesis testing, error metrics, equivalence testing, and top-set overlap analysis to provide a robust and quantifiable assessment of AI-expert agreement.


\item We showcase a case study rating design sketches on uniqueness, creativity, usefulness, and drawing quality, revealing that with carefully chosen methodology, one can coax near-expert judgments from a general VLM. 
\begin{itemize}[nosep]
    \item We show that top AI judges not only reach expert--level agreement for many metrics, but also meet expert agreement better or the same as trained novices across all metrics.
    \item  We confirm this \emph{quantitatively} by verifying equivalence with human experts using multi-pronged statistical checks. 
    
\end{itemize}

  \item We show that \textit{a handful of ratings} from experts can be scaled to \textit{rate thousands of designs} using in-context learning for VLMs while maintaining statistical equivalence to experts at a level higher than trained novices.



\end{itemize}

We emphasize that most of the statistical approaches themselves (TOST or correlation, hypothesis tests, etc.) are not our invention; rather, our contribution is to integrate these tests into a cohesive protocol, ensuring a thorough evaluation of AI-human alignment beyond single metrics and highlighting why they are necessary. This provides a template for other domains where researchers or practitioners might suspect that an AI is ``as good as'' a human expert, but need to validate that claim rigorously.

The remainder of the paper is organized as follows. In \textbf{Related Work}, we review established metrics for design evaluation, vision-language models in subjective evaluation tasks, in-context learning as a method of enhancing VLM-performance, and statistical methods for comparing AI with humans. The \textbf{Methodology} section describes our AI models, the experimental setup for training and evaluating them, the statistical procedure, and the decision criteria for expert equivalence. We then present \textbf{Results} from our experiments, including quantitative comparisons and illustrative figures. In \textbf{Discussion}, we interpret the findings, examine limitations, and outline future research directions. Finally, the \textbf{Conclusion} summarizes the contributions and significance of this study for the design engineering community.

\section{Related Works}
In the following sections, we describe related works on established metrics for evaluating engineering designs, the use of vision-language models for subjective evaluation, how in-context learning can be used to enhance VLMs even in data-scarce domains, and finally, established statistical methods for comparing raters.

\subsection{Engineering Design Metrics for Design Evaluation}
Engineering design literature has established a variety of metrics to evaluate the outcomes of conceptual design and ideation. A foundational distinction is that creative design quality is multi-dimensional: a concept must be both novel and functional to be truly valuable. 

Various metrics are used to measure the creativity of a design or idea. These metrics include, but are not limited to, expert panels~\cite{alipour2017impact,cheng2014new,chan2018best,baer2015importance,galati2015complexity}, the Consensual Assessment Technique (CAT)~\cite{amabile1982social,amabile1996creativity,baer2019assessing}, the Shah, Vargas-Hernandez and Smith (SVS) method~\cite{shah2003metrics}, and the Comparative Creativity Assessment (CCA), which is based on the SVS method~\cite{linsey2005collaborating}. Among all the metrics created, CAT~\cite{amabile1982social,amabile1996creativity,baer2019assessing} is often cited as the ``gold standard'' of creativity metrics; however, obtaining CAT ratings is very resource intensive~\cite{RN682,cseh2019scattered,kaufman2011expertise} as expert raters must code hundreds or thousands of ideas.

Many design studies employ expert judges to rate designs on Likert scales corresponding to such attributes. CAT uses human experts, averaging their scores to arrive at a reliable ground truth for creativity. CAT relies heavily on the intraclass correlation coefficient (ICC), and, particularly on having a sufficient ICC between experts~\cite{LeBreton2008}. ICC signifies ``the extent to which the mean rating assigned by a group of judges is reliable”\cite{LeBreton2008}, and typically ICC values of 0.7 or greater are considered acceptable levels of agreement among judges~\cite{James1984, Lance2006}. Although experts can be consistent, they are costly to obtain. Work by Miller et al., explores the use of trained human novices as proxies for experts~\cite{miller2021should}. They find that trained novices provide design ratings that are consistent with experts for novelty but less consistent for quality, and that novice agreement with experts varies if the metrics are social science or engineering metrics. 

Furthermore, some algorithmic metrics (like Shah’s approach for novelty or variety) have been proposed, but they sometimes fail to capture the subjective nuances that experts weigh. Data-driven approaches combine features from sketches with machine learning to predict CAT-style scores~\cite{edwards2021design, song2023aemultimodal}, bridging the gap between purely automated metrics and human judgment. Our work builds on these approaches, leveraging machine learning to replicate expert-labeled design metrics.

\subsection{Vision-Language Models in Subjective Evaluation}
VLMs have shown strong performance on tasks where images and text interplay, such as image captioning.
Prior design-centered work~\cite{edwards2021design, su2023multimodal, yuan2021multimodal, song2023aemultimodal, edwards2021form} often \emph{trains} a model on a large dataset of sketches labeled with, say, creativity scores, so that the model can predict them on new sketches. Our approach differs: we do not train or fine-tune the underlying large model; we rely instead on in-context learning, specifically few-shot prompting~\cite{brown2020fewshot}, in which we provide the VLM with a handful of demonstration examples that illustrate how a human might label a design. The model then infers how to rate a new sketch in a similar manner. This approach is attractive in design contexts where data is scarce, or experts are reluctant to label hundreds of examples.

Recent studies have explored pre-trained LLMs and VLMs as evaluators. LLM-Eval~\cite{lin2023llmevalunifiedmultidimensionalautomatic} and ROME~\cite{zhou2023rome} examine VLMs' reasoning abilities, particularly when faced with unconventional images requiring beyond-common-sense interpretation. Findings suggest that even state-of-the-art VLMs struggle with counterintuitive scenarios~\cite{zhou2023rome}, highlighting the need for deeper evaluation of their reliability in subjective assessments—further motivating our work.

LLMs have also been used as judges for subjective tasks. LLM-as-a-Judge~\cite{zheng2023llmasjudge} investigates whether LLMs can assess chatbot responses as accurately as humans in Chatbot Arena~\cite{chiang2024chatbotarenaopenplatform}, comparing their evaluations to MT-Bench human ratings~\cite{Bai_2024_mtbench}. Results suggest strong LLMs (e.g., GPT-4) can approximate human judgments with 80\% agreement, offering a scalable alternative to human evaluation. However, biases such as position bias (favoring the first response) and verbosity bias (favoring longer responses) remain concerns~\cite{zheng2023llmasjudge}, which could similarly affect design evaluations.

In a review of VLMs for engineering design tasks, researchers have tested GPT-4V’s ability to assess design similarity by replicating human-based experiments~\cite{picard2024conceptmanufacturingevaluatingvisionlanguage, ahmed2018triplets}. Results show GPT-4V matches or exceeds human performance in self-consistency and transitive reasoning.  Motivated by the opportunities of using VLMs as a judge, as well as the need to ensure that VLMs' ratings can match those of experts, our work explores the effectiveness of VLMs in subjective evaluation of designs. However, we move past just off-the-shelf VLMs and incorporate advancements in in-context learning to better align an AI judge with experts.
\begin{figure*}
    \centering
    \includegraphics[width=1\linewidth]{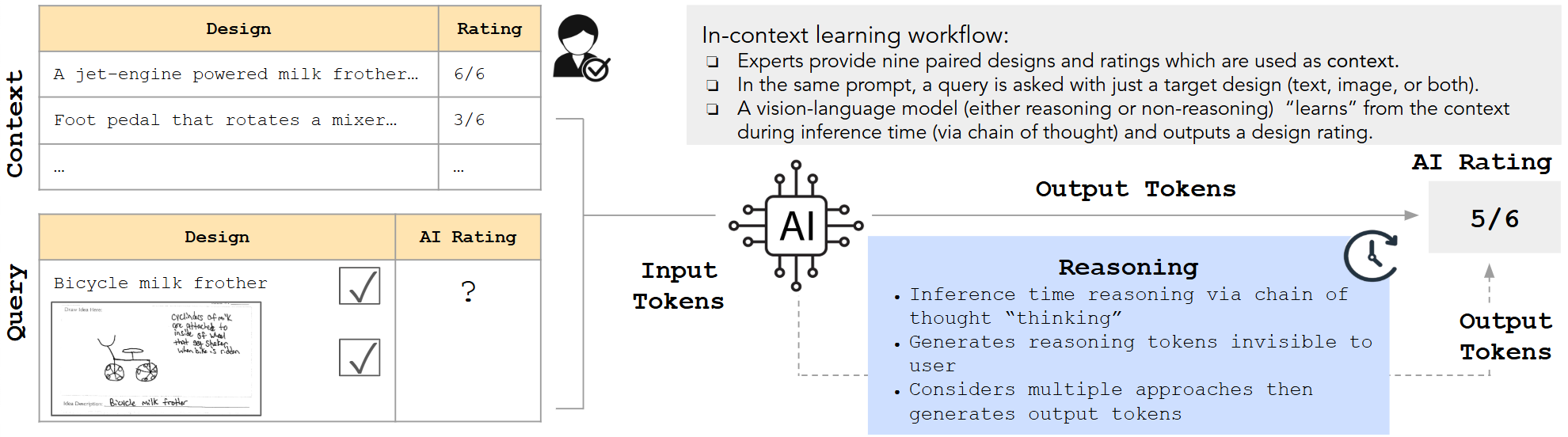}
    \caption{The in-context learning (ICL) workflow utilized to develop our four AI judges.}
    \label{fig:icl}
\end{figure*}

\subsection{In-Context Learning}
Few-shot in-context learning (ICL) is the ability of a large-language model (LLM) to perform a new task given a prompt that includes a few demonstrations of that task. The LLM, having seen correct input-output pairs in the demonstrations, can effectively produce a new output given a new input~\cite{brown2020fewshot, luo2024incontextlearningretrieveddemonstrations, dong-etal-2024-survey}. 

Thus, ICL is a way to adapt a language model to a new downstream task without fine-tuning or retraining a model, as was the traditional way of making a model work for a new task. A visual of how ICL is utilized for the AI judges in our case-study is shown in Figure~\ref{fig:icl}. ICL is particularly useful for data-scarce tasks. One does not need as much data for a new task as they would need if they were fine-tuning a model, since ICL only requires a few input-output pair demonstrations~\cite{luo2024incontextlearningretrieveddemonstrations, dong-etal-2024-survey}.  

Luo et al. identify several factors affecting ICL performance~\cite{luo2024incontextlearningretrieveddemonstrations}: the number and format of demonstrations, their order (with later examples possibly having a greater impact), and their diversity. Generally, more demonstrations benefit LLMs, although improvements diminish as the count grows, and the maximum context size remains a constraint.

\subsection{Comprehensive Statistical Methods for AI-Human Equivalence}
To evaluate AI model performance in comparison to human experts, we rely on established statistical methods. In this section, we introduce key statistical approaches relevant to our study, including measures of inter-rater reliability, absolute error metrics, rank correlation, and equivalence testing. While correlation or significance tests are common in evaluating an AI’s alignment with humans, they do not fully prove equivalence. We incorporate distribution checks (Friedman test), post-hoc Wilcoxon comparisons, correlation analysis, and TOST equivalence testing under a user-defined margin. 
Our novelty is applying all these tests \emph{alongside} other standard tests used in design to thoroughly confirm or refute ``human-level'' claims. By also leveraging measures like intraclass correlation and top-set overlap, one can see exactly where and how an AI’s judgments diverge from experts.

\textbf{Inter-Rater Reliability and Agreement}  
Reliable statistical methods are essential for assessing agreement between AI models and human experts on ordinal Likert-scale ratings. Cohen’s Kappa quantifies inter-rater agreement while accounting for chance~\cite{Cohen1960}, and its weighted variant penalizes larger disagreements, making it ideal for ordinal data~\cite{FleissCohen1973}. The Intraclass Correlation Coefficient (ICC) captures both correlation and absolute agreement, particularly useful when ratings are continuous and multiple raters are involved~\cite{ShroutFleiss1979}. Furthermore, ICC is a backbone metric for the Consensual Assessment Technique commonly used in design evaluation~\cite{LeBreton2008, miller2021should}.  

\textbf{Error Metrics for AI Performance}  
Absolute error metrics quantify how closely AI ratings align with human ratings. Mean Absolute Error (MAE) and Root Mean Square Error (RMSE) measure rating deviations, with RMSE penalizing larger errors more heavily~\cite{ChaiDraxler2014}. Given the nature of our ratings, ranging from 1-6 on a Likert scale, any differences will be small, making MAE a sufficient measurement of error.

Beyond these measures, Bland-Altman analysis visualizes agreement by plotting rating differences against their mean, revealing systematic biases that simple correlation cannot detect~\cite{BlandAltman1986}. This is particularly relevant in AI evaluation, where models may systematically over- or underestimate ratings. 

\textbf{Statistical Hypothesis Testing for Rating Differences}
To determine whether AI-generated ratings differ significantly from human expert ratings, we apply nonparametric statistical tests that do not assume normality in the data. The Friedman test, a nonparametric alternative to repeated-measures ANOVA, evaluates whether there are overall differences in ratings across multiple raters, making it well-suited for within-subjects experimental designs~\cite{Conover1999}. If significant differences are detected, post-hoc pairwise comparisons using the Wilcoxon signed-rank test can identify which AI models systematically differ from human experts~\cite{Wilcoxon1945}. This step is critical for determining whether AI-generated ratings fall within typical human rating variability.

\textbf{Correlation Analysis for Trend Agreement}
While absolute agreement metrics assess rating consistency, it is also important to evaluate whether AI models preserve the ranking of sketches relative to human assessments. Spearman’s rank correlation is a nonparametric measure that captures the monotonic relationship between two sets of rankings, making it suitable for ordinal data~\cite{Spearman1904}. A high Spearman correlation between AI and expert ratings suggests that even if absolute scores differ, the AI model successfully distinguishes between high- and low-quality designs in a manner consistent with human intuition.  

\textbf{Equivalence Testing for AI–Expert Comparability}  
We apply equivalence testing to assess whether AI ratings fall within a predefined human-expert tolerance range. The Two One-Sided Tests (TOST) procedure confirms practical AI-human parity, providing evidence beyond standard difference testing~\cite{Schuirmann1987}.  

\textbf{Top-Set Overlap Analysis for Design Selection}  
For applications where only the highest-rated designs are of interest, it is crucial to measure whether AI models select the same top-tier designs as human experts. The Jaccard similarity coefficient provides a metric for quantifying overlap between two sets, such as the top 10\% of sketches chosen by an AI model and an expert~\cite{Jaccard1901}. By computing Jaccard similarity across multiple top-set thresholds, we assess whether AI models prioritize the same high-quality designs as human evaluators, a key consideration in design selection tasks.

\section{VLM Methodology}

We developed four ICL approaches for rating design sketches. Unlike typical supervised learning that requires a large training set, we show each model only a \emph{few} demonstration examples. These include (1) a reference design sketch image, (2) a short textual description or prompt, and (3) the expert’s known rating for that design. The model thus observes several (e.g., 9) such examples in the prompt (this is the context), followed by a new target design (this is the query) and must generate a rating. No parameter updates occur; the large model remains unchanged. We simply rely on the model’s ability, via few-shot ICL, to mimic the rating process it saw in the examples.

\subsection{Overview of In-Context Models} 

As described above, and shown in Figure~\ref{fig:icl}, different types of information can be provided in both the context and the query. The value of using an AI judge is that the evaluation is more automated, thus it was critical that all of the information used in the context and query could be automatically retrieved. As such, we first used a vision-language model to extract the handwritten text descriptions that were on each sketch. Specifically, we instructed GPT-4o to extract the text descriptions from each design sketch via the API, and saved these descriptions as to only run this automated step once. This provided the textual description used in our later experiments. 

The four AI judges that we developed each receive a description of the design task, the design metric of interest (e.g. uniqueness), and the rating scale on which to provide the ratings. Unless otherwise stated, the VLM used was GPT-4o, specifically the ``gpt-4o-2024-08-06'' release. The AI judges' different contexts and queries are as follows:
\begin{enumerate}[nosep]
    \item \textbf{AI Judge: No Context} 
    \begin{enumerate}[nosep]
        \item Context: None. 
        \item Query: An image of the target design sketch.
    \end{enumerate}
    \item \textbf{AI Judge: Text} 
    \begin{enumerate}[nosep]
        \item Context: Nine textual descriptions of different designs and their expert ratings.
        \item Query: A textual description of the target design.
    \end{enumerate} 
    \item \textbf{AI Judge: Text + Image}
    \begin{enumerate}[nosep]
        \item Context: Nine textual descriptions of different designs and their expert ratings.
        \item Query: A textual description and an image of the target design sketch.
    \end{enumerate}
        \item \textbf{AI Judge: Text + Image + Reasoning}
    \begin{enumerate}[nosep]
        \item Context: Nine textual descriptions of different designs and their expert ratings.
        \item Query: A textual description and an image of the target design sketch.
        \item Model: OpenAI o1 reasoning model.
    \end{enumerate}
\end{enumerate}


Through initial experiments exploring the number and nature of context demonstrations, we found that using nine context designs along with a tenth target query provided a reliable framework. To minimize confounding variables, we kept these parameters constant across all judges and runs. While systematically varying these values is an important avenue for future research, it falls outside the scope of this work.


Furthermore, for each design metric, we selected nine context designs, which were provided to all AI judges except for the "No Context" judge. As detailed in Section~\ref{data_prep}, 50 images were set aside as a test set. In each of the three runs, and independently for each design metric, we selected nine context designs where both experts had agreed on their ratings. These designs were chosen to ensure a uniform distribution across the rating scale (1–6).


\section{Statistical Approach for Comparing AI Judges and Human Experts in Design Ratings}


We evaluated the four AI judges (AI Judge: No Context, AI Judge: Text, AI Judge: Text + Image, and AI Judge: Text + Image + Reasoning) against two human experts (Expert~1 and Expert~2) and 3 trained novices (Trained Novice 1-3) who rated a total of 934 early design sketches. This dataset\footnote{\url{https://sites.psu.edu/creativitymetrics/2018/07/18/milkfrother/}} originates from prior studies~\cite{starkey2019creativity, miller2016design, starkey2016abandoning}, where each sketch was evaluated on a 6-point Likert scale across four design metrics: \emph{uniqueness}, \emph{creativity}, \emph{usefulness}, and \emph{drawing quality}. These ratings were provided by two experts and three trained novices. The expert raters were classified as such based on their graduate degrees or completion of graduate coursework in an engineering design-related field\cite{miller2021should}. Our overarching goals were to determine: 
\begin{enumerate}[nosep]
    \item Which AI judge most closely aligns with the human experts' ratings, and
    \item Whether any AI judges perform at a level comparable to a \textit{human expert}, and 
    \item Whether any AI judges perform at a level comparable to a \textit{trained novice}.
\end{enumerate}

\subsection{Data Preparation}
\label{data_prep}
We reserved 50 sketches—where both experts provided matching ratings—for training the AI models. The full set of 934 sketches, including these training examples, was rated by two experts, three trained novices, and four AI judges. To ensure an unbiased evaluation, only the remaining 884 sketches were included in the test set. Finally, any sketches missing ratings from any judge (expert, trained novice, or AI) were removed, resulting in a final test set of approximately 875 sketches.

\subsection{Validation Metrics and Statistical Analyses}
To thoroughly assess each AI model's performance relative to the human experts, we combined multiple statistical approaches. These were selected based on the \emph{nature of the data} (ordinal Likert ratings, limited to discrete integer values from 1 to 6), the \emph{goal of measuring both absolute and relative agreement}, and the \emph{need for nonparametric methods} given non-normal distributions. All analyses were repeated separately for each of the five rating metrics.

\paragraph{Inter-Rater Agreement}  
We assess AI-human consistency using Weighted Cohen’s Kappa (quadratic) and Intraclass Correlation Coefficient (ICC). Kappa accounts for chance-corrected ordinal agreement, while ICC measures absolute agreement. Higher values indicate stronger reliability.

\vspace{0.5em}
\noindent \emph{Rationale:}  
Kappa is suited for ordinal Likert data, adjusting for partial agreement, while ICC treats ratings as continuous and reflects both correlation and agreement. Together, they ensure a robust assessment of AI performance relative to human experts.

\paragraph{Error Metrics (MAE)}  
Mean Absolute Error (MAE) quantifies rating deviations:
\begin{equation}
    \text{MAE} = \frac{1}{N} \sum_{i=1}^N \left|\, \hat{r}_i - r_i \right|
\end{equation}  
where $\hat{r}_i$ is the AI rating and $r_i$ is the expert rating. AI models are considered human-level if their MAE matches or is lower than expert-expert MAE.

\vspace{0.5em}
\noindent \emph{Rationale:}  
MAE directly measures rating accuracy and allows comparison against expert disagreement levels. Since there is no absolute ground truth, we compute MAE separately for each expert reference.

\paragraph{Bias and Consistency (Bland--Altman Analysis)}  
Bland--Altman analysis detects systematic bias by measuring mean rating differences and their variability. This demonstrates both the mean difference (or bias) between raters, as well as the standard deviation (SD) between the raters. If the bias is close to zero and most data points fall within $\pm1.96 \times$~SD of that difference, the two raters exhibit strong agreement. If AI--expert differences fall within the expert--expert limits of agreement (LoA), the AI demonstrates human-like rating consistency.

\vspace{0.5em}
\noindent \emph{Rationale:}  
Standard correlation does not always reveal systematic rating biases.  Bland--Altman analysis highlights potential biases and the spread of disagreements at different rating levels, thereby indicating if the AI's errors are consistent with normal human disagreement. Bland-Altman plots can reveal any systematic trends (heteroscedasticity). For example, perhaps the models agree with humans on low scores but diverge on very high scores. A trend line in the Bland-Altman (differences growing with the mean rating) would indicate the agreement depends on the sketch’s overall quality rating. 

\paragraph{Statistical Hypothesis Testing}  
We use the \textbf{Friedman test} to detect overall rating differences between experts, AI models, and trained novices. If significant ($p<0.05$), \textbf{Wilcoxon signed-rank tests} with Bonferroni correction identify specific AI-human differences.

\vspace{0.5em}
\noindent \emph{Rationale:}  
The Friedman test is non-parametric and suitable for repeated measures. Pairwise Wilcoxon tests pinpoint which AI models differ significantly from experts while controlling for multiple comparisons.

\paragraph{Rank Correlation (Spearman’s \texorpdfstring{$\rho$}{rho}))}  
Spearman’s rank correlation measures whether AI models maintain the relative ordering of sketches. A high AI--expert correlation near the expert--expert correlation suggests strong alignment in ranking.

\vspace{0.5em}
\noindent \emph{Rationale:}  
Even if absolute ratings differ, AI models should ideally rank sketches similarly to experts. Spearman’s $\rho$ evaluates this ranking agreement.

\paragraph{Equivalence Testing (TOST)}  
The \textbf{Two One-Sided Tests (TOST)} framework assesses whether AI ratings are statistically equivalent to expert ratings within a predefined tolerance (e.g., $\pm 1.0$ points). A significant result confirms AI-expert rating similarity.

\vspace{0.5em}
\noindent \emph{Rationale:}  
Non-significant differences (e.g., from Friedman or Wilcoxon tests) do not confirm equivalence; they only show \emph{no evidence of difference}. TOST directly tests whether any differences are \emph{small enough} to be practically irrelevant, reinforcing or contradicting findings of ``no difference.''

Equivalence testing complements hypothesis testing. A non-significant difference test means no detected difference, but a non-significant equivalence test means insufficient evidence of equivalence—often due to variability. If both tests support equivalence, there is good evidence that an AI judge is matching an expert. If difference is non-significant but equivalence fails, the AI is close but lacks definitive proof of expert-level performance.  



\paragraph{ Top-Set Overlap Analysis (Jaccard Similarity at Variable Cutoffs)} 

Another practical way to evaluate whether the AI models identify the same ``best'' sketches as the experts is to measure \emph{set overlap} in the top-rated designs. Although correlation and MAE capture broad patterns, some design use cases focus primarily on whether a model picks out the same top fraction of designs that a human would.

To compare an AI model and an expert at multiple thresholds, we define the following methodology: 
\begin{enumerate}[label=(\alph*)]

    \item \textbf{Nominal Fraction}. We start by choosing a series of nominal cutoffs (e.g., 5\%, 10\%, 15\%, \dots) to indicate the intended top $k\%$ of items for that expert's ratings. 
    
    \item \textbf{Reference Expert Top Set + Ties}. For each fraction $f$, let $N = \lceil f \cdot \text{(total sketches)} \rceil$. We collect the top $N$ sketches \emph{from the expert}, \emph{including all ties} at the boundary rating. In practice, if $N=10$ but the 10th and 11th sketches share the same rating, we include both. As a result, the final set may exceed $N$ items, yielding an \emph{actual fraction} larger than $f$. 
    
    \item \textbf{Model Top Set + Ties (Matching Size)}. Suppose the expert's top set ends up being $N_e$ sketches (after ties). We then form the model's top set by taking \emph{at least} $N_e$ items under the same tie rule. This ensures neither the expert set nor the model set excludes items arbitrarily cut off by rating ties. 
    
    \item \textbf{Jaccard Similarity}. Denote the expert's top set by $E_{\mathrm{top}}$ and the model's top set by $M_{\mathrm{top}}$. We measure \[ \mathrm{Jaccard}(E_{\mathrm{top}}, \; M_{\mathrm{top}}) \;=\; \frac{\bigl|E_{\mathrm{top}}\;\cap\;M_{\mathrm{top}}\bigr|} {\bigl|E_{\mathrm{top}}\;\cup\;M_{\mathrm{top}}\bigr|}. \] A higher Jaccard indicates better agreement on which sketches are considered best.
    
    \item \textbf{Plotting Actual Fraction vs.~Jaccard}. To compare the models and experts, we record the \emph{actual} fraction of total sketches $|E_{\mathrm{top}}| / (\text{total sketches})$ on the x-axis, since boundary ties can inflate (or occasionally reduce) the set size. On the y-axis, we plot the Jaccard similarity for each fraction. Doing this for multiple fractions (5\%, 10\%, 15\%, etc.) produces a curve showing how well the model and the expert align on which items belong in progressively larger top sets.

    \item \textbf{Comparing AUC of the Jaccard Plots} To quantitatively measure this metric, we find the AUC for the Jaccard plots for Expert 1 vs. Expert 2, and for all four AI judges vs. each expert as well as the three trained novices vs. each expert. Figures~\ref{fig:unique-jaccard}, \ref{fig:creativity-jaccard}, \ref{fig:useful-jaccard}, and~\ref{fig:drawing-jaccard} show these plots and the corresponding AUCs. 
\end{enumerate}

\noindent \emph{Rationale:}
Whereas correlation tests broad item-by-item rank alignment, top-set overlap directly answers, ``Does the model pick out the same top-tier sketches that the expert considers best?'' This is often crucial in design or idea-generation tasks where only the highest-performing concepts matter. By including ties at the threshold, we avoid arbitrary cutoffs (e.g., one design that shares the boundary rating gets left out of the model set but included in the expert set). Furthermore, comparing multiple fractions reveals if a model aligns strongly at small cutoffs (the absolute top 5\%) but diverges beyond that, or vice versa.

\subsection{Decision Criteria for ``Human-Level'' Ratings}
\label{statistical_criteria}
We deemed an AI judge ``as good as'' a human expert if, for a given metric:
\begin{enumerate}[nosep]

    \item \textbf{Agreement metric (Kappa):} The AI--expert Kappa value was within 20\% of the expert--expert baseline. Kappa measures categorical agreement beyond chance, and since higher values indicate stronger reliability, the AI--expert Kappa had to be at least $0.80 \times$ the expert--expert Kappa to meet this threshold.  

    \item \textbf{Agreement metric (ICC):} The AI--expert Intraclass Correlation Coefficient (ICC) was also within 20\% of the expert--expert ICC. Unlike Kappa, ICC quantifies the consistency of continuous \textit{or} ordinal ratings across raters. Kappa and ICC values were often equal. To ensure sufficient agreement, the AI--expert ICC needed to be at least $0.80 \times$ the expert--expert ICC. 
    
    \item \textbf{Error within expert range:} The AI's MAE against an expert did not exceed the expert--expert MAE within 20\%, indicating that average disagreements were no greater than typical human variability. So AI--expert MAE $\leq 1.2 \times$ expert--expert MAE.

    \item \textbf{Bland-Altman Bias:} The AI's mean difference, or bias, is within 20\% of the absolute mean bias between experts. 

    \item\textbf{Bland-Altman Limits of Agreement:} The AI's limits of agreement ($\pm 1.96 \times$ standard deviation of differences) are within 20\% of the expert--expert agreement bounds, ensuring AI predictions do not exhibit wider variability from an expert rater than another expert rater does. 
    
    \item \textbf{Equivalence (TOST):} The AI--expert differences lay fully within the $\pm1$ margin, reinforcing that any small discrepancies are practically negligible.
    
    \item \textbf{No significant distribution differences:} The Friedman test and paired Wilcoxon signed-rank tests showed no systematic bias or distributional shifts in AI vs. expert ratings  (i.e., Bonferroni-corrected $p>0.05$). 
    
    \item \textbf{Correlation near expert--expert:} Spearman's $\rho$ between AI and expert $\geq 0.8 \times$ the experts' own inter-correlation, indicating comparable ranking ability. Mathematically, 
    \[\rho_{AI-expert} \geq 0.8  \times \rho_{expert-expert} \]

    \item \textbf{Top-set overlap (Jaccard approach):} For various nominal cutoffs (e.g.\ 5\%, 10\%, 15\%), the AI’s \emph{actual} top set had a high Jaccard similarity with the expert’s top set across all those cutoffs. We compared the AI--expert overlap to the expert--expert overlap at each cutoff. 
    For the overall plot, we compared the area under the curve (AUC):
    \[
       AUC_{\text{Jaccard(AI, Expert)}} 
       \;\;\ge\;\; 0.8 \times AUC_{\text{Jaccard(Expert1, Expert2)}}
    \]
\end{enumerate}

We repeated these analyses independently for each of the four metrics (uniqueness, creativity, usefulness, and drawing quality).
We then summarized which judge(s) met which criteria, and thereby concluded which judge performed better and whether it operated at a ``expert-level'' for any or all of the metrics.
 It is possible an AI judge is very good on some metrics but not others, or beyond a trained-novice level but not yet at an expert level. We note such cases in the Results section and discuss overarching trends in the Discussion.

\vspace{1em}
\noindent
\emph{Practical Interpretation:} If an AI consistently demonstrates (1) close agreement metrics, (2) MAE $\simeq$ expert--expert MAE, (3) low bias and similar limits of agreement as expert--expert pairs, (4) a TOST equivalence outcome, and (5) no significant difference in distribution, (6) a strong correlation with the expert, then it can be treated as being ``as reliable as'' a human rater on that metric. Otherwise, we can examine where it fell short (e.g., higher error, lower correlation, or failing to show equivalence) to pinpoint which aspects of performance need improvement. Lastly, we compared the four AI judges in two respects: 1) among each other to determine which ICL setup yielded ratings that best matched experts, and 2) we compared the AI judges against the three trained human novices on the same criteria to identify if a well-developed AI judge shown just nine design-rating examples could equally or more closely match expert ratings than trained humans.

        \begin{table*}[!ht]\centering
            \caption{Run 1: Summary of Uniqueness results. Kappa and ICC are relative to the expert--expert baseline of 0.54. MAE values compare to the observed Expert--Expert MAE = 1.10. ``Equiv?'' indicates whether TOST found AI--expert equivalence within $\pm1.0$ margin. Bold indicates judge is within 20\% of expert-expert agreement.}
            \label{tab:uniqueness_run1}
            \begin{tabular}{ll c ccc cccc c}
            \toprule
              & \textbf{Comparison} & \textbf{Kappa} & \textbf{ICC} & \textbf{MAE} & \textbf{\shortstack{Mean Diff \\ $\pm$ 1.96 SD}} & \textbf{Equiv?} & \textbf{Spear.} & \textbf{\shortstack{Wilcoxon \\ p-corr}} & \textbf{AUC} & \textbf{\shortstack{Tests\\ Passed}} \\
            \midrule
            & \textbf{Expert 1 vs Expert 2} & 0.54 & 0.54 & 1.10 & $0.33 \pm 2.85$ & -- & 0.54 & $ 2.11 \cdot 10^{-09 }$ & 0.64 & -- \\
            \midrule
        \midrule\multirow{7}{*}{\rotatebox{90}{\textbf{Expert 1}}}
& Trained Novice 1 & \textbf{0.49} & \textbf{0.49} & \textbf{1.17} & $\textbf{-0.31} \pm \textbf{3.01}$ & \textbf{True} & \textbf{0.48} & $ 1.41 \cdot 10^{-08 }$ & \textbf{0.63} & 8/9 \\ 
& Trained Novice 2 & \textbf{0.44} & \textbf{0.44} & \textbf{1.20} & $0.83 \pm \textbf{2.79}$ & \textbf{True} & \textbf{0.50} & $ 6.54 \cdot 10^{-50 }$ & \textbf{0.63} & 7/9 \\ 
& Trained Novice 3 & \textbf{0.51} & \textbf{0.51} & \textbf{1.04} & $\textbf{-0.10} \pm \textbf{2.67}$ & \textbf{True} & \textbf{0.50} & $\mathbf{1.22 \cdot 10^{-01}}$ & \textbf{0.64} & 9/9 \\ 
& AI Judge: No Context & 0.24 & 0.24 & 1.58 & $-1.33 \pm \textbf{2.75}$ & False & \textbf{0.43} & $ 3.17 \cdot 10^{-92 }$ & \textbf{0.61} & 3/9 \\ 
& AI Judge: Text & \textbf{0.47} & \textbf{0.47} & \textbf{1.10} & $\textbf{-0.35} \pm \textbf{2.75}$ & \textbf{True} & \textbf{0.48} & $ 8.91 \cdot 10^{-12 }$ & \textbf{0.63} & 8/9 \\ 
& AI Judge: Text + Image & 0.42 & 0.42 & \textbf{1.15} & $\textbf{0.23} \pm \textbf{2.99}$ & \textbf{True} & 0.40 & $ 3.38 \cdot 10^{-05 }$ & \textbf{0.61} & 5/9 \\ 
& AI Judge: T + I + Reasoning & \textbf{0.49} & \textbf{0.49} & \textbf{1.08} & $\textbf{-0.38} \pm \textbf{2.66}$ & \textbf{True} & \textbf{0.52} & $ 8.00 \cdot 10^{-15 }$ & \textbf{0.64} & 8/9 \\ 
\midrule\multirow{7}{*}{\rotatebox{90}{\textbf{Expert 2}}}
& Trained Novice 1 & \textbf{0.58} & \textbf{0.58} & \textbf{1.07} & $\textbf{0.01} \pm \textbf{2.82}$ & \textbf{True} & \textbf{0.56} & $\mathbf{1.00 \cdot 10^{+00}}$ & \textbf{0.65} & 9/9 \\ 
& Trained Novice 2 & 0.42 & 0.42 & 1.40 & $1.15 \pm \textbf{2.77}$ & False & \textbf{0.53} & $ 1.73 \cdot 10^{-77 }$ & \textbf{0.66} & 3/9 \\ 
& Trained Novice 3 & \textbf{0.55} & \textbf{0.55} & \textbf{1.00} & $\textbf{0.23} \pm \textbf{2.62}$ & \textbf{True} & \textbf{0.55} & $ 1.88 \cdot 10^{-06 }$ & \textbf{0.65} & 8/9 \\ 
& AI Judge: No Context & 0.28 & 0.28 & 1.42 & $-1.00 \pm \textbf{2.89}$ & False & \textbf{0.45} & $ 1.34 \cdot 10^{-63 }$ & \textbf{0.63} & 3/9 \\ 
& AI Judge: Text & \textbf{0.53} & \textbf{0.53} & \textbf{1.06} & $\textbf{-0.02} \pm \textbf{2.71}$ & \textbf{True} & \textbf{0.54} & $\mathbf{1.00 \cdot 10^{+00}}$ & \textbf{0.64} & 9/9 \\ 
& AI Judge: Text + Image & 0.42 & 0.42 & \textbf{1.19} & $0.55 \pm \textbf{2.98}$ & \textbf{True} & \textbf{0.44} & $ 2.81 \cdot 10^{-23 }$ & \textbf{0.63} & 5/9 \\ 
& AI Judge: T + I + Reasoning & \textbf{0.52} & \textbf{0.52} & \textbf{1.04} & $\textbf{-0.05} \pm \textbf{2.71}$ & \textbf{True} & \textbf{0.54} & $\mathbf{3.54 \cdot 10^{-01}}$ & \textbf{0.64} & 9/9 \\ 
\bottomrule
\end{tabular}
\end{table*}

\section{Results}
\label{sec:results}
In this section, we report our findings for each metric (\emph{uniqueness}, \emph{creativity}, \emph{usefulness}, and \emph{drawing quality}). We compare the ratings from the four AI judges (No Context, Text, Text + Image, and Text + Image + Reasoning) against the ratings from the two human experts (Expert~1 and Expert~2). The expert--expert levels of agreement serve as a baseline. For each metric we provide the comprehensive results from one of our three runs, as shown in Tables~\ref{tab:uniqueness_run1}, \ref{tab:creativity_run1}, \ref{tab:usefulness_run1}, and \ref{tab:drawing_run1}, while the results for runs two and three are provided in the Appendix. We report statistical measures of agreement, error, hypothesis tests (Friedman and post-hoc Wilcoxon), and equivalence (TOST). For clarity, the nine statistical tests that we report in Tables~\ref{tab:uniqueness_run1}-\ref{tab:creativity_run3} are outlined in Section~\ref{statistical_criteria}. Furthermore, for each metric we plot the Jaccard similarity between the Expert 1 and all other raters. 

\begin{figure}[htbp]
    \centering
    \includegraphics[width=1\linewidth]{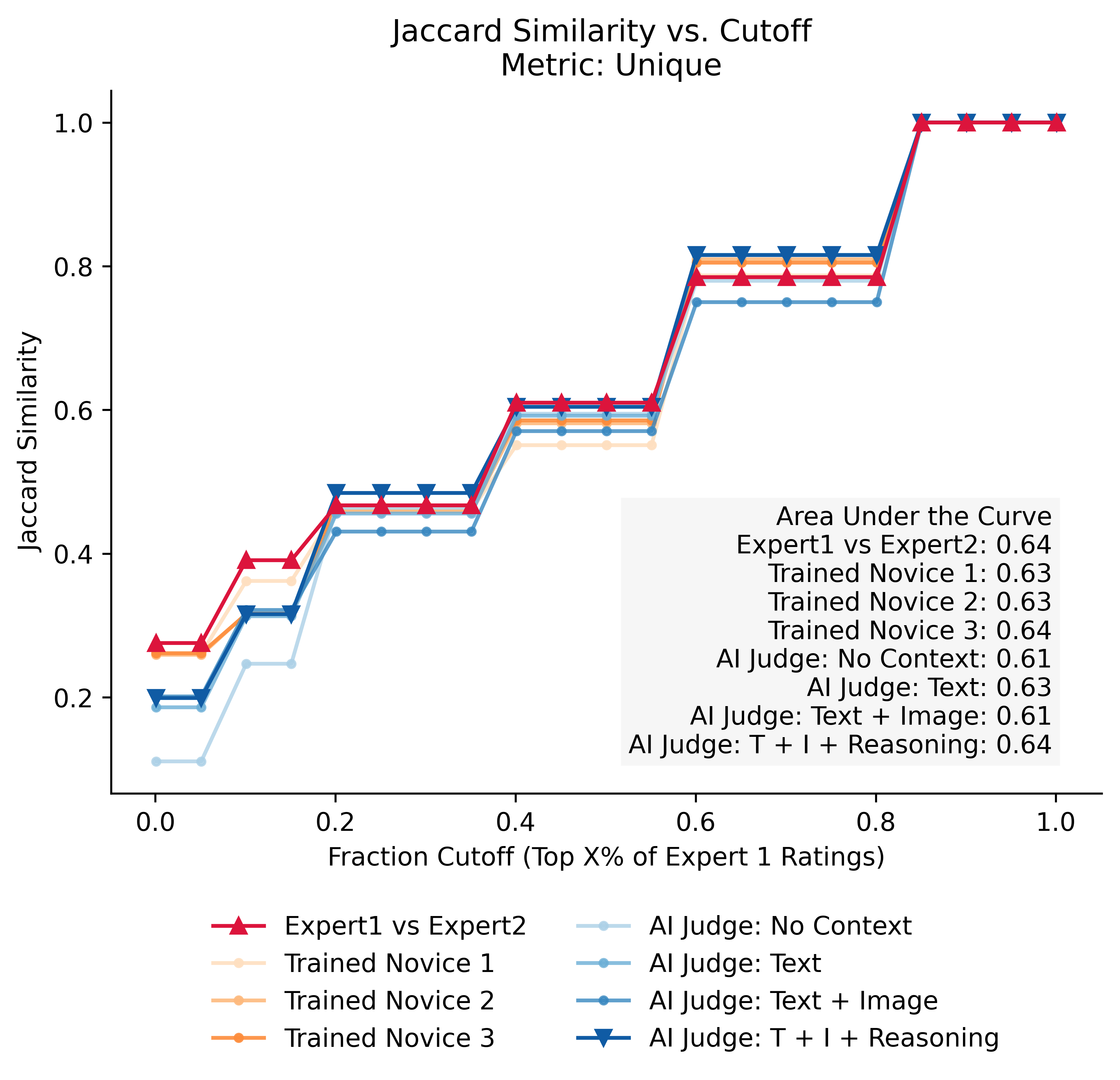}
    \caption{AI Judge: Text + Image + Reasoning reaches expert-level AUC, as does one trained novice, all other models do not.}
    \label{fig:unique-jaccard}
\end{figure}

\subsection{Uniqueness}

\begin{table}[!ht]\centering
\small  
    \caption{Uniqueness: Summary of Tests Passed Across Runs}
    \label{tab:unique_tests_passed}
    \begin{tabular}{p{0.1cm} p{3cm} cccc} 
    \toprule
      & \textbf{Comparison} & Run 1 & Run 2 & Run 3 & \textbf{Avg.} \\
    \midrule 
    \multirow{7}{*}{\rotatebox{90}{\textbf{Expert 1}}} 
    & \textit{Trained Novice 1} & 8/9 & 8/9 & 8/9 & 8/9\\
    & \textit{Trained Novice 2 }& 7/9 & 7/9 & 7/9 & 7/9\\
    & \textit{Trained Novice 3 }& 9/9 & 9/9 & 9/9 & 9/9\\
    & AI Judge: No Context & 3/9 & 2/9 & 2/9 & 2.33/9\\
    & AI Judge: Text & 8/9 & 6/9 & 5/9 & 6.33/9 \\
    & AI Judge: Text + Img. & 5/9 & 8/9 & 7/9 & 6.67/9 \\
    & AI Judge: Text + Img. + Reasoning. & 8/9 & 8/9 & 8/9 & 8/9\\
    
    \midrule 
    \multirow{7}{*}{\rotatebox{90}{\textbf{Expert 2}}} 
    & T\textit{rained Novice 1} & 9/9 & 9/9 & 9/9 & 9/9 \\
    & \textit{Trained Novice 2} & 3/9 & 3/9 & 3/9 & 3/9 \\
    & \textit{Trained Novice 3} & 8/9 & 8/9 & 8/9 & 8/9 \\
    & AI Judge: No Context & 3/9 & 2/9 & 3/9 & 2.67/9 \\
    & AI Judge: Text & 9/9 & 4/9 & 8/9 & 8.67/9 \\
    & AI Judge: Text + Img. & 5/9 & 9/9 & 8/9 & 7.33/9 \\
    & AI Judge: Text + Img. + Reasoning & 9/9 & 8/9 & 9/9 & 8.67/9 \\
    \bottomrule
\end{tabular}
\end{table}

\begin{figure*}[hbtp]
    \centering
    \begin{subfigure}{0.47\linewidth}
        \centering
        \includegraphics[width=\linewidth]{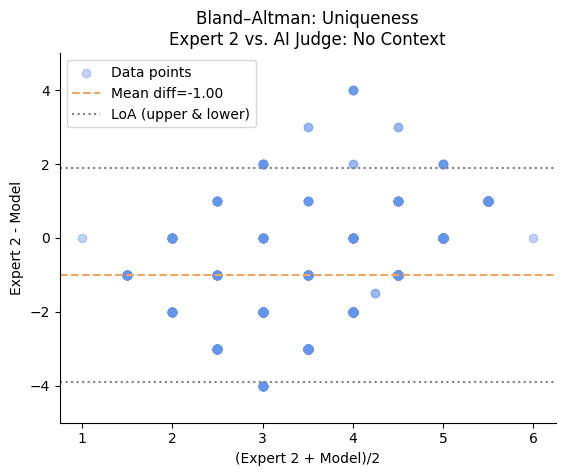}
        \caption{}
        \label{fig:bland_unique_no_context}
    \end{subfigure}
    \hfill
    \begin{subfigure}{0.47\linewidth}
        \centering
        \includegraphics[width=\linewidth]{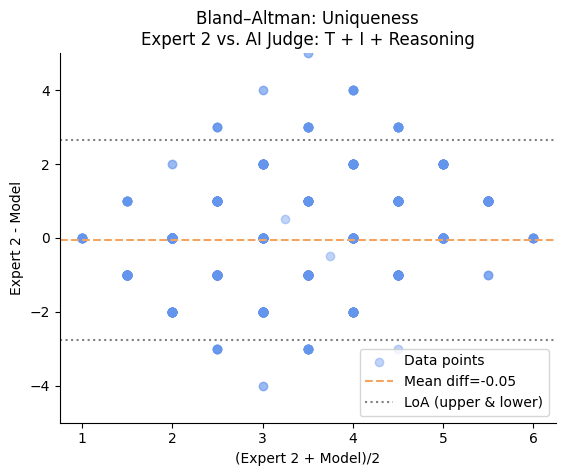}
        \caption{}
        \label{fig:bland-altman-unique-reasoning}
    \end{subfigure}
    \caption{Bland-Altman plots comparing AI judges' ratings to expert ratings. (a) The AI Judge with No Context consistently assigns higher uniqueness ratings than Expert 2. (b) The AI Judge with Reasoning shows minimal bias. Both instances reveal larger differences in the middle range of the mean ratings.}
    \label{fig:bland-altman-combined}
\end{figure*}

For rating uniqueness, the best-performing AI rater was AI Judge: Text + Image + Reasoning, which passed at least 8/9 statistical tests across all runs (Table \ref{tab:unique_tests_passed}). This model achieved agreement with experts within 20\% of expert–expert levels for all key metrics except the Wilcoxon post hoc test, where its p-value exceeded 0.05 (Table~\ref{tab:uniqueness_run1}). This performance (matching expert--expert agreement in at least 8/9 tests) is on par with the top two trained novices (who each pass one more test than this AI judge across the six runs), and outperforms one of the trained novices. Figure~\ref{fig:unique-jaccard} plots the Jaccard similarity between raters and Expert 1 at different top design percentiles. The graph shows that by a fraction cutoff of approximately 0.25, the AI Judge: Text + Image + Reasoning ratings have a Jaccard similarity with Expert 1 that is higher than or on par with the Jaccard similarity between the two experts for the remainder of the plot.

AI Judge: Text + Image showed moderate performance, averaging 6.67/9 tests passed against Expert 1 and 7.33/9 against Expert 2 (Table~\ref{tab:unique_tests_passed}). The only difference between this judge and the top-performing one is the inclusion of a reasoning model, suggesting that reasoning contributes meaningfully to expert-level agreement even when using identical context.

AI Judge: No Context consistently performed the worst for uniqueness, passing only 2–3 tests per run. This result underscores the importance of in-context learning for aligning with expert judgments. The performance gap is also evident in the Bland–Altman plots (Figures~\ref{fig:bland-altman-unique-reasoning} and~\ref{fig:bland_unique_no_context}): the reasoning model shows negligible systematic bias relative to the expert (mean difference near zero), whereas the No Context judge overestimates scores by about 1.0 point on average.

Lastly, we note some trends among which statistical tests are more often passed or failed when rating uniqueness. All raters, AI judges and trained novices, pass the TOST test, meaning their average rating differences stay within $\pm1.0$. All raters also pass the Jaccard similarity plot's AUC test. Many raters, similarly are within 20\% of expert-level agreement for MAE. Results for inter-rater agreement analysis (Kappa and ICC), as well as ranked correlation analysis (Spearman) are similar, where many, but not all raters fall within 20\% of expert-level agreement. The Wilcoxon p-correlation is a consistently difficult test for raters to reach expert-level agreement.

\subsection{Creativity}

        \begin{table*}[!htbp]\centering
            \caption{Run 1: Statistical test results for Creativity. Kappa and ICC are relative to the expert--expert baseline of 0.26. MAE values compare to the observed Expert--Expert MAE = 1.25. ``Equiv?'' indicates whether TOST found AI--expert equivalence within $\pm1.0$ margin. Bold indicates judge is within 20\% of expert-expert agreement.}
            \label{tab:creativity_run1}
            \begin{tabular}{ll c ccc cccc c}
            \toprule
              & \textbf{Comparison} & \textbf{Kappa} & \textbf{ICC} & \textbf{MAE} & \textbf{\shortstack{Mean Diff \\ $\pm$ 1.96 SD}} & \textbf{Equiv?} & \textbf{Spear.} & \textbf{\shortstack{Wilcoxon \\ p-corr}} & \textbf{\shortstack{Jacc. \\ AUC}} & \textbf{\shortstack{Tests\\ Passed}} \\
            \midrule
            & \textbf{Expert 1 vs Expert 2} & 0.26 & 0.26 & 1.25 & $0.19 \pm 3.13$ & -- & 0.22 & $ 2.39 \cdot 10^{-03 }$ & 0.59 & -- \\
            \midrule
        \midrule\multirow{7}{*}{\rotatebox{90}{\textbf{Expert 1}}}
& Trained Novice 1 & -0.02 & -0.02 & 1.62 & $-0.88 \pm \textbf{3.59}$ & \textbf{True} & -0.04 & $ 2.06 \cdot 10^{-37 }$ & \textbf{0.55} & 3/9 \\ 
& Trained Novice 2 & 0.08 & 0.09 & \textbf{1.38} & $0.60 \pm \textbf{3.39}$ & \textbf{True} & 0.11 & $ 5.69 \cdot 10^{-23 }$ & \textbf{0.57} & 4/9 \\ 
& Trained Novice 3 & -0.07 & -0.07 & 1.50 & $-0.58 \pm \textbf{3.62}$ & \textbf{True} & -0.07 & $ 4.22 \cdot 10^{-17 }$ & \textbf{0.57} & 3/9 \\ 
& AI Judge: No Context & 0.05 & 0.05 & 2.04 & $-1.89 \pm \textbf{2.97}$ & False & 0.11 & $ 1.68 \cdot 10^{-113 }$ & \textbf{0.58} & 2/9 \\ 
& AI Judge: Text & 0.11 & 0.11 & \textbf{1.48} & $-1.02 \pm \textbf{3.00}$ & False & 0.15 & $ 7.29 \cdot 10^{-61 }$ & \textbf{0.58} & 3/9 \\ 
& AI Judge: Text + Image & 0.11 & 0.11 & \textbf{1.40} & $-0.56 \pm \textbf{3.24}$ & \textbf{True} & 0.11 & $ 8.20 \cdot 10^{-21 }$ & \textbf{0.59} & 4/9 \\ 
& AI Judge: T + I + Reasoning & 0.15 & 0.15 & \textbf{1.37} & $-0.91 \pm \textbf{2.84}$ & \textbf{True} & \textbf{0.21} & $ 1.98 \cdot 10^{-55 }$ & \textbf{0.59} & 5/9 \\ 
\midrule\multirow{7}{*}{\rotatebox{90}{\textbf{Expert 2}}}
& Trained Novice 1 & 0.17 & 0.17 & \textbf{1.32} & $-0.70 \pm \textbf{3.03}$ & \textbf{True} & \textbf{0.20} & $ 2.34 \cdot 10^{-33 }$ & \textbf{0.60} & 5/9 \\ 
& Trained Novice 2 & 0.15 & 0.15 & \textbf{1.34} & $0.79 \pm \textbf{3.07}$ & \textbf{True} & \textbf{0.18} & $ 1.20 \cdot 10^{-40 }$ & \textbf{0.60} & 5/9 \\ 
& Trained Novice 3 & \textbf{0.23} & \textbf{0.23} & \textbf{1.20} & $-0.39 \pm \textbf{2.91}$ & \textbf{True} & \textbf{0.25} & $ 1.29 \cdot 10^{-12 }$ & \textbf{0.59} & 7/9 \\ 
& AI Judge: No Context & 0.09 & 0.09 & 1.83 & $-1.70 \pm \textbf{2.65}$ & False & \textbf{0.22} & $ 1.19 \cdot 10^{-114 }$ & \textbf{0.58} & 3/9 \\ 
& AI Judge: Text & 0.20 & 0.20 & \textbf{1.26} & $-0.83 \pm \textbf{2.66}$ & \textbf{True} & \textbf{0.26} & $ 4.57 \cdot 10^{-53 }$ & \textbf{0.58} & 5/9 \\ 
& AI Judge: Text + Image & 0.16 & 0.16 & \textbf{1.20} & $-0.37 \pm \textbf{2.99}$ & \textbf{True} & 0.17 & $ 1.86 \cdot 10^{-12 }$ & \textbf{0.58} & 4/9 \\ 
& AI Judge: T + I + Reasoning & \textbf{0.22} & \textbf{0.22} & \textbf{1.17} & $-0.73 \pm \textbf{2.56}$ & \textbf{True} & \textbf{0.29} & $ 5.48 \cdot 10^{-45 }$ & \textbf{0.60} & 7/9 \\ 
\bottomrule
\end{tabular}
\end{table*}

\begin{table}[!ht]\centering
\small  
    \caption{Creativity: Summary of Statistical Tests Passed Across Runs}
    \label{tab:creativity_tests_passed}
    \begin{tabular}{p{0.1cm} p{3cm} cccc} 
    \toprule
      & \textbf{Comparison} & Run 1 & Run 2 & Run 3 & \textbf{Avg.} \\
    \midrule 
    \multirow{7}{*}{\rotatebox{90}{\textbf{Expert 1}}} 
    & \textit{Trained Novice 1} & 3/9 & 3/9 & 3/9 & 3/9\\
    & \textit{Trained Novice 2} & 4/9 & 4/9 & 4/9 & 4/9\\
    & \textit{Trained Novice 3} & 3/9 & 3/9 & 4/9 & 3.33/9\\
    & AI Judge: No Context & 2/9 & 2/9 & 2/9 & 2/9\\
    & AI Judge: Text & 3/9 & 4/9 & 4/9 & 3.67/9 \\
    & AI Judge: Text + Img. & 4/9 & 2/9 & 2/9 & 2.67/9 \\
    & AI Judge: Text + Img. + Reason. & 5/9 & 3/9 & 4/9 & 4/9\\
    
    \midrule 
    \multirow{7}{*}{\rotatebox{90}{\textbf{Expert 2}}} 
    & \textit{Trained Novice 1} & 5/9 & 5/9 & 5/9 & 5/9 \\
    & \textit{Trained Novice 2} & 5/9 & 5/9 & 5/9 & 5/9 \\
    & \textit{Trained Novice 3} & 7/9 & 7/9 & 7/9 & 7/9 \\
    & AI Judge: No Context & 3/9 & 3/9 & 3/9 & 3/9 \\
    & AI Judge: Text & 5/9 & 4/9 & 4/9 & 4.33/9 \\
    & AI Judge: Text + Img. & 4/9 & 5/9 & 4/9 & 4.33/9 \\
     & AI Judge: Text + Img. + Reason. & 7/9 & 7/9 & 5/9 & 6.33/9 \\
    \bottomrule
\end{tabular}
\end{table}


\begin{figure}[!htbp]
    \centering
    \includegraphics[width=1\linewidth]{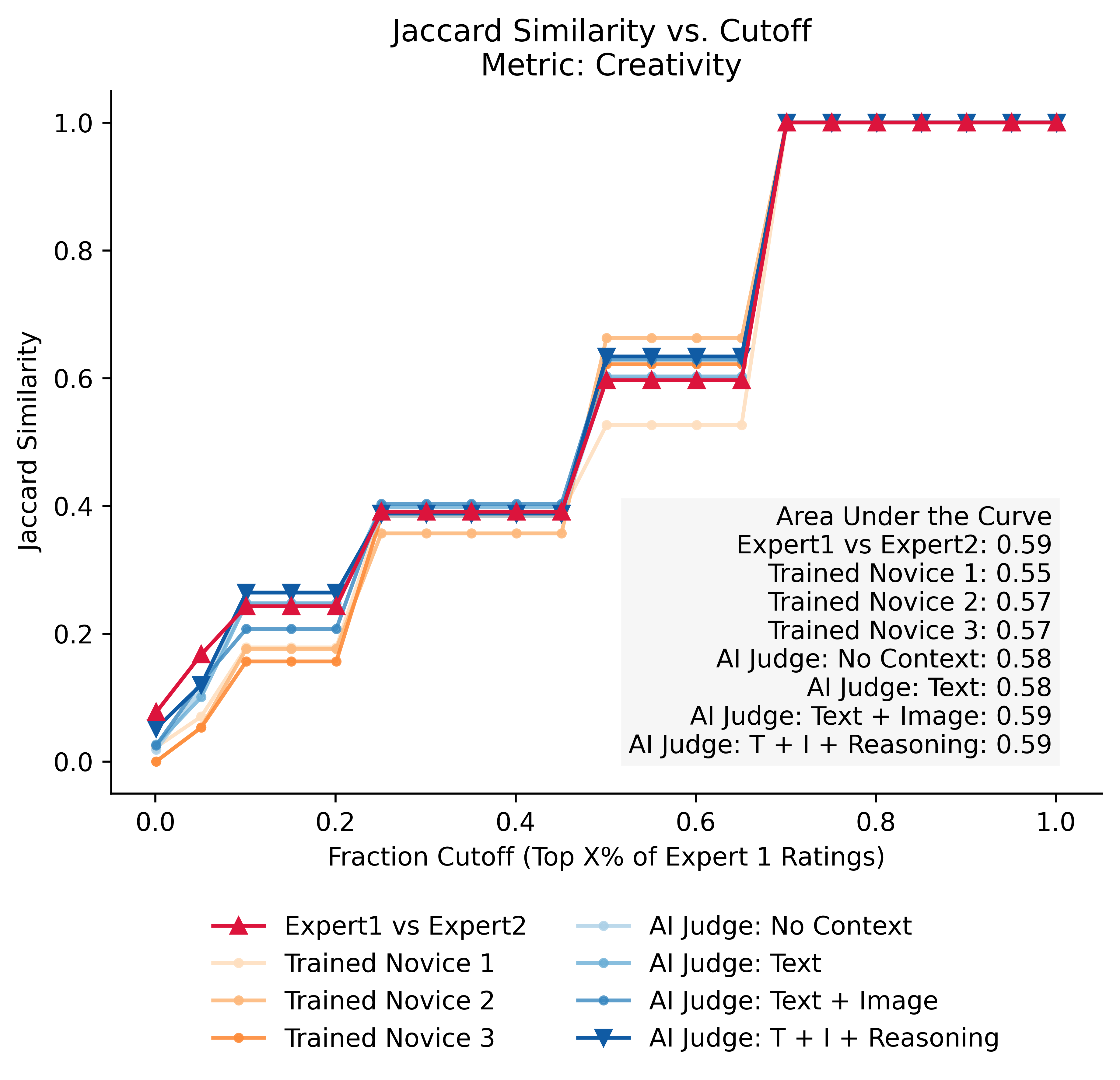}
    \caption{Two of the AI judges, shown in blue, reach the same AUC as Expert 1 vs. Expert 2, while the trained novices do not.}
    \label{fig:creativity-jaccard}
\end{figure}

Experts, AI judges, and trained novices all exhibited lower statistical similarity when rating creativity compared to uniqueness in this study. Expert–expert Kappa, ICC, Spearman, and Jaccard similarity AUC were lower for creativity, while mean average error was higher. This aligns with prior studies using the same dataset, which independently found creativity to be the most challenging metric for AI to predict~\cite{edwards2021design, song2023aemultimodal}.

Among AI judges, AI Judge: Text + Image + Reasoning performed best, as detailed in Table~\ref{tab:creativity_run1} and summarized across runs in Table~\ref{tab:creativity_tests_passed}. This model consistently outperformed two of the three trained novices against both experts, with only one trained novice surpassing it in a single instance (Expert 2, run three).

Non-reasoning AI judges performed on par with or worse than trained novices, with AI Judge: No Context ranking the lowest. This mirrors the uniqueness results and reinforces the hypothesis that few-shot ICL significantly improves vision-language model performance. However, unlike in uniqueness ratings, AI Judge: Text (text-only input) performed as well as or better than AI Judge: Text + Image across experts and runs.

\subsection{Usefulness}

        \begin{table*}[!htbp]\centering
            \caption{Run 1: Statistical test results for Usefulness. Kappa and ICC are relative to the expert--expert baseline of 0.59. MAE values compare to the observed Expert--Expert MAE = 1.00. ``Equiv?'' indicates whether TOST found AI--expert equivalence within $\pm1.0$ margin. Bold indicates judge is within 20\% of expert-expert agreement.}
            \label{tab:usefulness_run1}
            \begin{tabular}{ll c ccc cccc c}
            \toprule
              & \textbf{Comparison} & \textbf{Kappa} & \textbf{ICC} & \textbf{MAE} & \textbf{\shortstack{Mean Diff \\ $\pm$ 1.96 SD}} & \textbf{Equiv?} & \textbf{Spear.} & \textbf{\shortstack{Wilcoxon \\ p-corr}} & \textbf{\shortstack{Jacc. \\ AUC}} & \textbf{\shortstack{Tests\\ Passed}} \\
            \midrule
            & \textbf{Expert 1 vs Expert 2} & 0.59 & 0.60 & 1.00 & $-0.01 \pm 2.61$ & -- & 0.58 & $\mathbf{1.00 \cdot 10^{+00}}$ & 0.67 & -- \\
            \midrule
        \midrule\multirow{7}{*}{\rotatebox{90}{\textbf{Expert 1}}}
& Trained Novice 1 & 0.12 & 0.12 & 1.53 & $-0.81 \pm 1.67$ & \textbf{True} & 0.15 & $ 4.41 \cdot 10^{-36 }$ & \textbf{0.55} & 3/9 \\ 
& Trained Novice 2 & 0.34 & 0.35 & 1.60 & $-1.21 \pm 1.65$ & False & 0.45 & $ 4.77 \cdot 10^{-68 }$ & \textbf{0.62} & 2/9 \\ 
& Trained Novice 3 & 0.35 & 0.36 & 1.20 & $-0.36 \pm \textbf{2.92}$ & \textbf{True} & 0.36 & $ 1.37 \cdot 10^{-11 }$ & \textbf{0.60} & 4/9 \\ 
& AI Judge: No Context & 0.29 & 0.29 & 1.37 & $-0.91 \pm \textbf{2.75}$ & \textbf{True} & 0.37 & $ 1.50 \cdot 10^{-57 }$ & \textbf{0.61} & 4/9 \\ 
& AI Judge: Text & 0.35 & 0.35 & 1.27 & $-0.50 \pm \textbf{3.01}$ & \textbf{True} & 0.38 & $ 7.63 \cdot 10^{-19 }$ & \textbf{0.61} & 4/9 \\ 
& AI Judge: Text + Image & 0.29 & 0.29 & 1.33 & $-0.34 \pm 1.64$ & \textbf{True} & 0.30 & $ 5.37 \cdot 10^{-09 }$ & \textbf{0.59} & 3/9 \\ 
& AI Judge: T + I + Reasoning & 0.35 & 0.35 & 1.30 & $-0.77 \pm \textbf{2.81}$ & \textbf{True} & 0.42 & $ 5.79 \cdot 10^{-42 }$ & \textbf{0.62} & 4/9 \\ 
\midrule\multirow{7}{*}{\rotatebox{90}{\textbf{Expert 2}}}
& Trained Novice 1 & 0.15 & 0.15 & 1.54 & $-0.83 \pm 1.70$ & \textbf{True} & 0.17 & $ 3.93 \cdot 10^{-37 }$ & \textbf{0.54} & 3/9 \\ 
& Trained Novice 2 & 0.33 & 0.33 & 1.71 & $-1.22 \pm 1.72$ & False & 0.43 & $ 1.91 \cdot 10^{-64 }$ & \textbf{0.60} & 2/9 \\ 
& Trained Novice 3 & 0.24 & 0.25 & 1.38 & $-0.37 \pm 1.66$ & \textbf{True} & 0.22 & $ 1.70 \cdot 10^{-09 }$ & \textbf{0.56} & 3/9 \\ 
& AI Judge: No Context & 0.16 & 0.17 & 1.51 & $-0.92 \pm 1.62$ & False & 0.20 & $ 9.10 \cdot 10^{-48 }$ & \textbf{0.56} & 2/9 \\ 
& AI Judge: Text & 0.20 & 0.21 & 1.47 & $-0.52 \pm 1.76$ & \textbf{True} & 0.20 & $ 5.01 \cdot 10^{-16 }$ & \textbf{0.56} & 3/9 \\ 
& AI Judge: Text + Image & 0.14 & 0.15 & 1.54 & $-0.36 \pm 1.85$ & \textbf{True} & 0.13 & $ 3.25 \cdot 10^{-07 }$ & \textbf{0.54} & 3/9 \\ 
& AI Judge: T + I + Reasoning & 0.19 & 0.20 & 1.51 & $-0.78 \pm 1.68$ & \textbf{True} & 0.22 & $ 7.94 \cdot 10^{-35 }$ & \textbf{0.55} & 3/9 \\ 
\bottomrule
\end{tabular}
\end{table*}

\begin{table}[!ht]\centering
\small  
    \caption{Usefulness: Summary of Tests Passed Across Runs}
    \label{tab:usefulness_tests_passed}
    \begin{tabular}{p{0.1cm} p{3cm} cccc} 
    \toprule
      & \textbf{Comparison} & Run 1 & Run 2 & Run 3 & \textbf{Avg.} \\
    \midrule 
    \multirow{7}{*}{\rotatebox{90}{\textbf{Expert 1}}} 
    & \textit{Trained Novice 1} & 3/9 & 3/9 & 3/9 & 3/9\\
    & \textit{Trained Novice 2 }& 2/9 & 2/9 & 2/9 & 2/9\\
    & \textit{Trained Novice 3} & 4/9 & 4/9 & 5/9 & 4.33/9\\
    & AI Judge: No Context & 4/9 & 4/9 & 4/9 & 4/9\\
    & AI Judge: Text & 4/9 & 3/9 & 4/9 & 3.67/9 \\
    & AI Judge: Text + Img. & 3/9 & 3/9 & 4/9 & 3.33/9 \\
    & AI Judge: Text + Img. + Reason. & 4/9 & 4/9 & 3/9 & 3.67/9\\
    
    \midrule 
    \multirow{7}{*}{\rotatebox{90}{\textbf{Expert 2}}} 
    & \textit{Trained Novice 1} & 3/9 & 3/9 & 3/9 & 3/9 \\
    & \textit{Trained Novice 2} & 2/9 & 2/9 & 2/9 & 2/9 \\
    & \textit{Trained Novice 3} & 3/9 & 3/9 & 3/9 & 3/9 \\
    & AI Judge: No Context & 2/9 & 3/9 & 3/9 & 2.67/9 \\
    & AI Judge: Text & 3/9 & 2/9 & 3/9 & 2.67/9 \\
    & AI Judge: Text + Img. & 3/9 & 2/9 & 3/9 & 2.67/9 \\
    & AI Judge: Text + Img. + Reason. & 3/9 & 3/9 & 2/9 & 2.67/9 \\
    \bottomrule
\end{tabular}
\end{table}

\begin{figure}[htbp]
    \centering
    \includegraphics[width=1\linewidth]{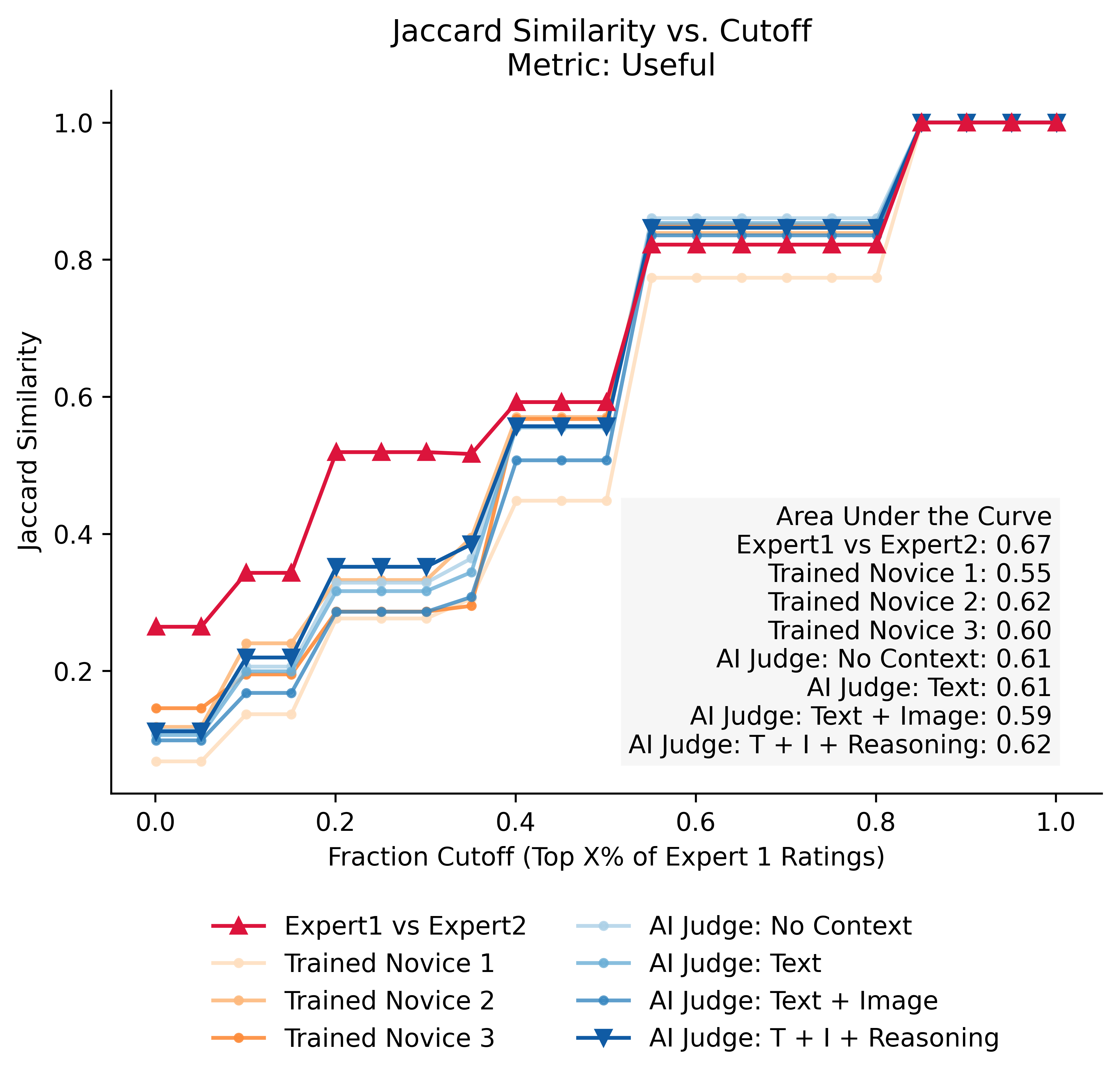}
    \caption{AI Judge: Text + Image + Reasoning and Trained Novice 2 come closest to expert-level AUC of 0.67, with their AUC of 0.62. In general, for Usefulness, expert-level Jaccard similarity is hard for novices and AI judges to meet.}
    \label{fig:useful-jaccard}
\end{figure}
 
No single AI judge consistently outperformed the others in assessing usefulness. AI Judge: No Context performed better than expected, passing 4 out of 9 statistical tests in most runs against Expert 1—outperforming two trained novices (Table~\ref{tab:usefulness_tests_passed}). Against Expert 2, it performed on par with the other AI judges, all averaging just 2.67/9 tests passed. AI Judge: Text and AI Judge: Text + Image showed similar performance, passing 2–4 tests per run. Surprisingly, AI Judge: Text + Image + Reasoning did not show a clear advantage, averaging 3.67/9 tests against Expert 1 and 2.67/9 against Expert 2. This suggests that reasoning contributed less to aligning with expert usefulness ratings (Table~\ref{tab:usefulness_tests_passed}).

Among trained novices, Trained Novice 3 performed best, averaging 4.33/9 tests against Expert 1 and 3/9 against Expert 2, slightly outperforming AI judges. Trained Novices 1 and 2 performed worse, passing 3/9 and 2/9 tests, respectively, highlighting the general difficulty in achieving expert-level agreement on usefulness.

Both AI judges and trained novices performed best in TOST equivalence and Jaccard similarity plot AUC, but struggled with inter-rater agreement. Kappa and ICC values remained well below the expert-expert baseline (0.59), with most AI judges scoring between 0.14 and 0.35. Notably, usefulness was the only metric where expert ratings were significantly similar in the Wilcoxon post hoc test (p > 0.05). However, no AI judge or trained novice passed this test, indicating systematic differences in their rating distributions.

\begin{figure}[htbp]
    \centering
    \includegraphics[width=1\linewidth]{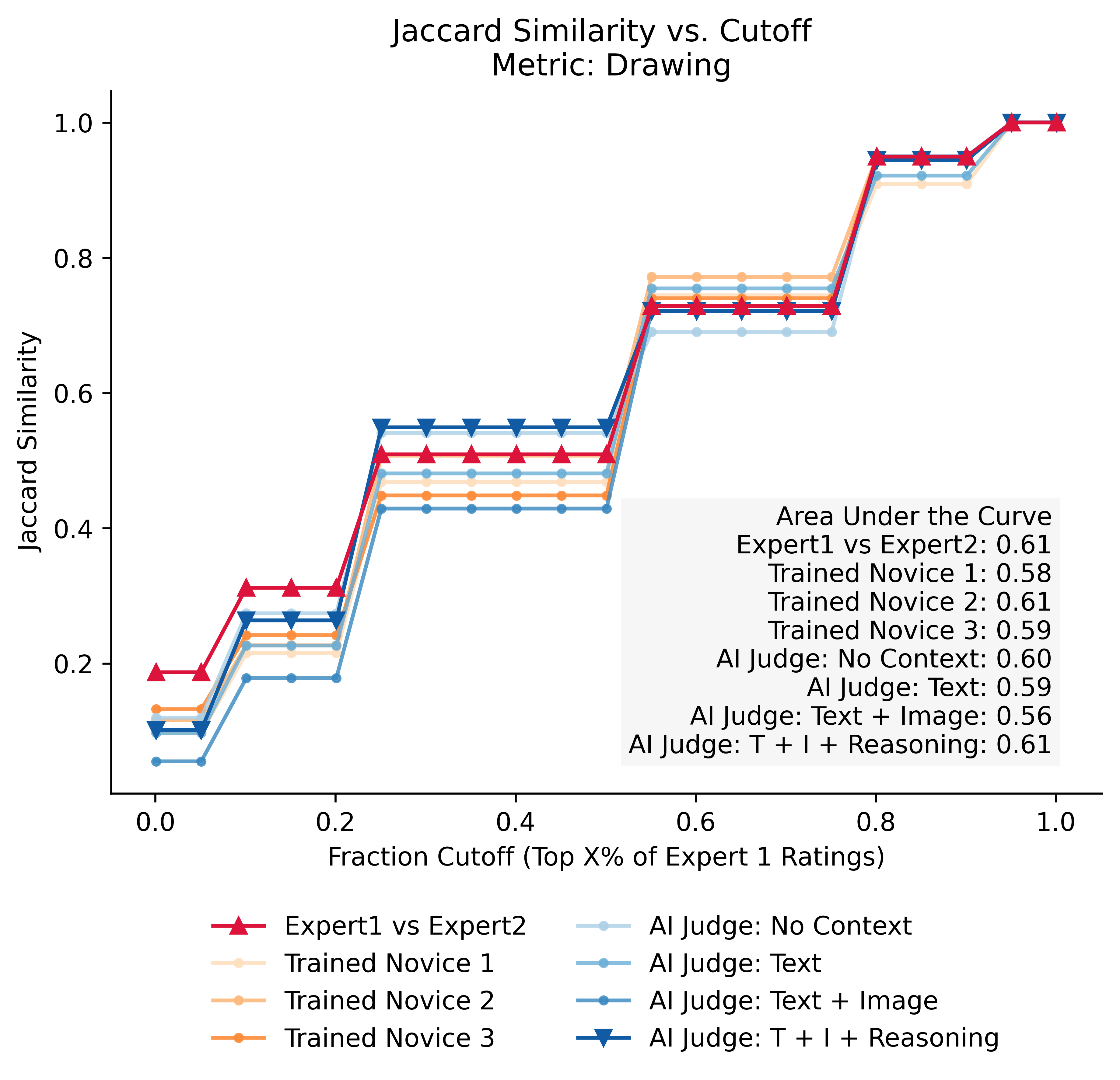}
    \caption{For Drawing Quality, AI Judge: Text + Image + Reasoning and Trained Novice 2 reach expert-level agreement for Jaccard similarity (AUC = 0.61). Interestingly, AI Judge: No Context also reaches a high AUC of 0.60, higher than the other two trained novices.}
    \label{fig:drawing-jaccard}
\end{figure}

The Jaccard similarity plot for usefulness (Figure~\ref{fig:useful-jaccard}) further highlights this challenge—all raters fail to reach expert-expert similarity levels until a much higher fraction cutoff than for other metrics (Figures~\ref{fig:unique-jaccard}, \ref{fig:creativity-jaccard}, and\ref{fig:drawing-jaccard}). This suggests that while experts largely agree on usefulness ratings, neither AI judges nor trained novices can reliably match their assessments.




\subsection{Drawing Quality}

        \begin{table*}[htbp]\centering
            \caption{Run 1: Statistical test results for Drawing Quality. Kappa and ICC are relative to the expert--expert baseline of 0.33. MAE values compare to the observed Expert--Expert MAE = 1.16. ``Equiv?'' indicates whether TOST found AI--expert equivalence within $\pm1.0$ margin. Bold indicates judge is within 20\% of expert-expert agreement.}
            \label{tab:drawing_run1}
            \begin{tabular}{ll c ccc cccc c}
            \toprule
              & \textbf{Comparison} & \textbf{Kappa} & \textbf{ICC} & \textbf{MAE} & \textbf{\shortstack{Mean Diff \\ $\pm$ 1.96 SD}} & \textbf{Equiv?} & \textbf{Spear.} & \textbf{\shortstack{Wilcoxon \\ p-corr}} & \textbf{\shortstack{Jacc. \\ AUC}} & \textbf{\shortstack{Tests\\ Passed}} \\
            \midrule
            & \textbf{Expert 1 vs Expert 2} & 0.33 & 0.33 & 1.16 & $-0.67 \pm 2.65$ & -- & 0.37 & $ 2.19 \cdot 10^{-39 }$ & 0.61 & -- \\
            \midrule
        \midrule\multirow{7}{*}{\rotatebox{90}{\textbf{Expert 1}}}
& Trained Novice 1 & 0.17 & 0.17 & \textbf{1.26} & $\textbf{0.44} \pm \textbf{3.00}$ & \textbf{True} & 0.17 & $ 1.13 \cdot 10^{-15 }$ & \textbf{0.58} & 5/9 \\ 
& Trained Novice 2 & 0.24 & 0.24 & 1.45 & $0.92 \pm \textbf{3.07}$ & False & 0.27 & $ 4.57 \cdot 10^{-49 }$ & \textbf{0.61} & 2/9 \\ 
& Trained Novice 3 & 0.22 & 0.22 & \textbf{1.23} & $0.81 \pm \textbf{2.64}$ & \textbf{True} & 0.25 & $ 3.31 \cdot 10^{-52 }$ & \textbf{0.59} & 4/9 \\ 
& AI Judge: No Context & \textbf{0.28} & \textbf{0.28} & \textbf{1.02} & $\textbf{0.40} \pm \textbf{2.42}$ & \textbf{True} & \textbf{0.32} & $ 4.44 \cdot 10^{-19 }$ & \textbf{0.60} & 8/9 \\ 
& AI Judge: Text & 0.20 & 0.20 & \textbf{1.19} & $\textbf{0.74} \pm \textbf{2.63}$ & \textbf{True} & 0.26 & $ 8.14 \cdot 10^{-46 }$ & \textbf{0.59} & 5/9 \\ 
& AI Judge: Text + Image & 0.13 & 0.14 & \textbf{1.30} & $\textbf{0.78} \pm \textbf{2.91}$ & \textbf{True} & 0.17 & $ 1.85 \cdot 10^{-43 }$ & \textbf{0.56} & 5/9 \\ 
& AI Judge: T + I + Reasoning & \textbf{0.31} & \textbf{0.31} & \textbf{0.99} & $\textbf{0.22} \pm \textbf{2.47}$ & \textbf{True} & \textbf{0.32} & $ 1.06 \cdot 10^{-06 }$ & \textbf{0.61} & 8/9 \\ 
\midrule\multirow{7}{*}{\rotatebox{90}{\textbf{Expert 2}}}
& Trained Novice 1 & \textbf{0.37} & \textbf{0.37} & \textbf{0.98} & $\textbf{-0.23} \pm \textbf{2.55}$ & \textbf{True} & \textbf{0.36} & $ 1.70 \cdot 10^{-06 }$ & \textbf{0.62} & 8/9 \\ 
& Trained Novice 2 & \textbf{0.45} & \textbf{0.45} & \textbf{1.02} & $\textbf{0.25} \pm \textbf{2.64}$ & \textbf{True} & \textbf{0.46} & $ 2.36 \cdot 10^{-07 }$ & \textbf{0.66} & 8/9 \\ 
& Trained Novice 3 & \textbf{0.43} & \textbf{0.43} & \textbf{0.88} & $\textbf{0.14} \pm \textbf{2.27}$ & \textbf{True} & \textbf{0.42} & $ 2.57 \cdot 10^{-03 }$ & \textbf{0.63} & 8/9 \\ 
& AI Judge: No Context & \textbf{0.38} & \textbf{0.38} & \textbf{0.85} & $\textbf{-0.27} \pm \textbf{2.17}$ & \textbf{True} & \textbf{0.42} & $ 6.16 \cdot 10^{-12 }$ & \textbf{0.65} & 8/9 \\ 
& AI Judge: Text & \textbf{0.27} & \textbf{0.27} & \textbf{0.96} & $\textbf{0.06} \pm \textbf{2.51}$ & \textbf{True} & 0.28 & $\mathbf{4.80 \cdot 10^{-01}}$ & \textbf{0.62} & 8/9 \\ 
& AI Judge: Text + Image & 0.11 & 0.11 & \textbf{1.15} & $\textbf{0.11} \pm \textbf{2.92}$ & \textbf{True} & 0.12 & $\mathbf{1.73 \cdot 10^{-01}}$ & \textbf{0.58} & 6/9 \\ 
& AI Judge: T + I + Reasoning & \textbf{0.29} & \textbf{0.29} & \textbf{0.99} & $\textbf{-0.46} \pm \textbf{2.38}$ & \textbf{True} & \textbf{0.33} & $ 2.22 \cdot 10^{-25 }$ & \textbf{0.63} & 8/9 \\ 
\bottomrule
\end{tabular}
\end{table*}

    

\begin{table}[!ht]\centering
\small  
    \caption{Drawing Quality: Summary of Tests Passed Across Runs}
    \label{tab:drawing_quality_tests_passed}
    \begin{tabular}{p{0.1cm} p{3cm} cccc} 
    \toprule
      & \textbf{Comparison} & Run 1 & Run 2 & Run 3 & \textbf{Avg.} \\
    \midrule 
    \multirow{7}{*}{\rotatebox{90}{\textbf{Expert 1}}} 
    & \textit{Trained Novice 1} & 5/9 & 5/9 & 5/9 & 5/9\\
    & \textit{Trained Novice 2} & 2/9 & 2/9 & 2/9 & 2/9\\
    & \textit{Trained Novice 3} & 4/9 & 4/9 & 4/9 & 4/9\\
    & AI Judge: No Context & 8/9 & 8/9 & 8/9 & 8/9\\
    & AI Judge: Text & 5/9 & 5/9 & 6/9 & 5.33/9 \\
    & AI Judge: Text + Img. & 5/9 & 5/9 & 8/9 & 6/9 \\
    & AI Judge: Text + Img. + Reason. & 8/9 & 8/9 & 9/9 & 8.33/9\\
    
    \midrule 
    \multirow{7}{*}{\rotatebox{90}{\textbf{Expert 2}}} 
    & \textit{Trained Novice 1} & 8/9 & 8/9 & 8/9 & 8/9 \\
    &\textit{ Trained Novice 2} & 8/9 & 8/9 & 8/9 & 8/9 \\
    & \textit{Trained Novice 3} & 8/9 & 8/9 & 8/9 & 8/9 \\
    & AI Judge: No Context & 8/9 & 8/9 & 8/9 & 8/9 \\
    & AI Judge: Text & 8/9 & 6/9 & 5/9 & 6.33/9 \\
    & AI Judge: Text + Img. & 6/9 & 5/9 & 5/9 & 5.33/9 \\
    & AI Judge: Text + Img. + Reason. & 8/9 & 8/9 & 6/9 & 7.33/9 \\
    \bottomrule
\end{tabular}
\end{table}

The AI Judge with No Context and AI Judge with Text + Image + Reasoning were the top-performing AI judges for drawing quality ratings. AI Judge: No Context passed 8/9 statistical tests across all runs, while AI Judge: Text + Image + Reasoning passed 8/9 statistical tests in all runs but one. The strong performance of AI Judge: No Context suggests that raw visual assessment alone is highly effective for evaluating drawing quality. This finding is supported by prior findings in~\cite{song2023aemultimodal}. Meanwhile, AI Judge: Text + Image + Reasoning demonstrated consistently strong agreement with expert ratings, reinforcing the benefits of multimodal reasoning.

Both of these AI judges had superior overall performance to the trained novices, which showed high rating agreement with Expert 2 (8/9 tests passed across the board), but low rating agreement with Expert 1 (an average of 2/9, 4/9, and 5/9 tests passed across three runs - Tables~\ref{tab:drawing_run1},~\ref{tab:drawing_run2}, and~\ref{tab:drawing_run3}). 

Across the three runs, the metrics for which the AI judges are best able to meet expert--expert agreement levels are MAE, Bland-Altman's mean difference and limits of agreement, TOST equivalence,and Jaccard similarity plot's AUC. The two lesser-performing AI judges, AI Judge: Text and AI Judge: Text + image struggle with inter-rater agreement metrics, Kappa and AUC, as well as the Spearman correlation coefficient.

That said, drawing quality is one of the design metrics where the best AI judges are able to meet expert--expert agreement levels consistently. Interesting, while the trained novices perform relatively poorly compared to Expert 1, they are able to exceed expert--expert agreement with Expert 2 for the metrics of Kappa, ICC, MAE, and Spearman correlation coefficient (Tables~\ref{tab:drawing_run1},~\ref{tab:drawing_run2}, and~\ref{tab:drawing_run3}). AI Judge: No Context is similarly able to exceed expert--expert agreement when compared with Expert 2 for these metrics.

\section{Discussion}

The results from each of the four design metrics have been explored in Section~\ref{sec:results}. The AI Judge: Text + Image + Reasoning has better overall performance than two of the three trained novices for all four metrics, as measured by the total number of tests passed across all three runs and compared to each expert. Most often, the top two raters with best agreement with experts are AI Judge: Text + Image + Reasoning and Trained Novice 3. Both of these raters frequently pass 8/9 statistical tests, and in some cases 9/9 statistical tests, indicating that they are able to match expert ratings with agreement levels on par with fellow experts. 

\textbf{Comparison of AI Judges}
Among the four AI judges we developed, AI Judge: Text + Image + Reasoning best matched expert--expert agreement levels for uniqueness, creativity, and (tied) drawing quality. This supports the notion that reasoning models equipped with ICL may more effectively replicate expert ratings than their non-reasoning counterparts. In contrast, AI Judge: No Context showed the weakest alignment with expert-level agreement for uniqueness and creativity, highlighting the value of few-shot demonstrations via ICL.

This trend is particularly evident in Tables~\ref{tab:unique_tests_passed} and~\ref{tab:creativity_tests_passed}, where the three ICL-based judges consistently outperformed the No Context model in aligning with expert ratings for uniqueness and creativity. Interestingly, however, AI Judge: No Context outperformed many of the others for usefulness and drawing quality. For drawing quality, this may be because the model was still given the target image in the query—suggesting that the VLM’s pretrained understanding of visual quality may suffice without demonstrations. This aligns with findings from~\cite{song2023aemultimodal}, which reported that drawing quality was the only design metric where image input improved AI predictions more than text descriptions.

In some cases, AI Judge: Text performed as well as—or even better than—AI Judge: Text + Image, particularly for creativity, usefulness, and in Expert 2 comparisons for uniqueness and drawing quality. This result is somewhat counterintuitive, as one might expect that additional contextual information (i.e., the image) would enhance performance. One possible explanation is that the experts may have relied more heavily on the textual descriptions than the sketches when making their assessments. This aligns with findings from~\cite{song2023aemultimodal}, which showed that text descriptions had greater predictive power than sketches. Future work could further investigate this phenomenon.

\textbf{Statistical Tests often Passed or Failed} In general passing the Wilcoxon post hoc test is the rarest, while TOST equivalence and Jaccard similarity plot AUC are the most common. We selected $\pm1.0$ as our margin of equivalence for TOST. The choice is grounded in the idea that an AI doesn’t need to be perfect, just within normal expert variation. Based on a given application and desired strictness of matching, a smaller TOST equivalence margin could be used, and perhaps fewer cases would declare equivalence.

Another notable trend among the statistical tests is that of Bland-Altman bias. This is measured for each run as (expert rating - model rating). For uniqueness, creativity, and usefulness, the bias between either expert and any AI judge is consistently negative. This means that the AI judge almost \textit{always} rates a design higher (if only slightly) than an expert. For trained novices the same pattern holds, although not for every single run. The only design metric where this is not the case is drawing quality. Here, the bias is usually positive when compared to Expert 1, (Expert 1's ratings > model's ratings) and often negative when compared to Expert 2 (Expert 2's ratings < model's ratings). This is very useful for determining if and how an AI's ratings can complement or supplement an expert's ratings.

\textbf{Trends Among Design Metrics} Given that this is just one case study, we cannot make sweeping claims about what metrics can best be predicted by the AI judges, but we do note trends found in our results. Between Expert 1 and Expert 2, statistical agreement measures like Kappa, ICC, and Spearman's $\rho$ are higher for uniqueness and usefulness. However, trained novices and AI judges alike, are better able to agree with experts for uniqueness and drawing quality and actually show the \textit{lowest agreement} for usefulness (Figure~\ref{fig:useful-jaccard} and Table~\ref{tab:usefulness_tests_passed}). This trend is corroborated with past work that has used AI models to predict expert ratings from this same dataset, and found that experts' usefulness and creativity ratings are harder to predict than their uniqueness and drawing quality ratings~\cite{song2023aemultimodal}.

\textbf{Implications for In-Context VLMs in Design}
Our results show that \emph{large, general vision-language models} can approach expert-level performance on subjective design metrics with minimal overhead, provided that we carefully craft the prompting. This is a powerful finding for design scenarios with limited data or budget to collect massive labels for training. The key is choosing demonstration examples that illustrate the rating scale well, so the model sees how humans scored certain reference sketches.

AI judges' strong performance when compared to trained novices has promising cost and time saving implications for design evaluations. These findings also suggest that AI judges could be used in place of trained novices as proxies for expert raters. As a showcase of the potential savings, let us assume an expert is paid \$200/hour to rate designs based on uniqueness. Each design takes the expert 15 seconds to rate, so 1,000 designs would cost $\$200 \times 4.167 \text{ hours}=\$833.33$. With the ICL framework, an expert could rate just 9 designs, totaling \$12.50. From those 9 designs, we can use a reasoning model to rate the remainder of the 1,000. In our 12 trials with OpenAI's o1, running the reasoning model cost $\sim$ \$0.17/ design, so for 991 additional designs, it would cost \$168.47. Therefore, the total cost savings are $\$833.33 - \$12.50 -\$168.47 = \$652.46$, or \textbf{78\% cost savings}. 

This value becomes even clearer when considering the cost and time of hiring experts and the opportunity cost of using their time for repetitive ratings. Unlike human raters, who experience rating fatigue and shifts in judgment~\cite{ling2014fatigue}, AI provides consistent evaluations at scale. Lastly, the cost of LLMs and VLMs is steadily decreasing\cite{korinek2023cheaper, epoch2025llminferencepricetrends}, especially with advancements in open-source models\cite{liu2023visualinstructiontuning}, making AI-assisted evaluation increasingly viable.

\textbf{Statistical Rigor Matters} We emphasize that it is insufficient to rely on a single metric (e.g., correlation or a p-value from a difference test) to claim human-level performance. Our evaluation integrates multiple statistical approaches to provide a comprehensive assessment of AI judge reliability. Each statistical test captures a different aspect of agreement; some focus on absolute error, while others assess ranking or distributional similarity. Passing one test does not necessarily imply passing others, highlighting the need for a multifaceted evaluation framework.

To exemplify this, for the drawing quality metric, Trained Novice 3 frequently passed the TOST equivalence test and exhibited low MAE when compared to Expert 1, yet failed in weighted Cohen’s Kappa, ICC, and Spearman correlation. This suggests that while their ratings were numerically close on average, their ranking patterns and relative consistency differed significantly. Such discrepancies highlight the limitations of single-metric evaluations and motivate our use of a suite of nine statistical tests to capture different facets of expert equivalence. By incorporating multiple measures, we ensure a more rigorous assessment of whether a model or rater truly aligns with expert judgment across both absolute agreement and ranking consistency.


Our approach also emphasized selecting appropriate parameters and thresholds for the design context. For example, $\pm1.0$ equivalence margin in our TOST analysis and the 20\% error margin from expert–expert agreement were determined by the Likert scale used and expert agreement in our dataset. Future work may explore ways to determine the best parameters for a given application. This flexibility allows for tailoring AI evaluation to different levels of precision required in design assessment tasks.

\section{Limitations and Future Work}
Though encouraging, these findings should be tested on other design tasks (e.g., CAD models, design prototypes) to confirm generality. Because in-context learning can fail if the domain is too far from the model’s pretraining, we might not always see such strong results. Another limitation is interpretability: the AI’s rating might match an expert’s but not necessarily provide a rationale. Future efforts can study the “chain-of-thought” text so the model explains how it judged certain design aspects. Nonetheless, the synergy between prompt engineering and robust validation sets a precedent for deploying AI in design critiques.

We plan to further study whether the trends among design metrics hold for other datasets. Are uniqueness and drawing quality consistently the metrics with which AI judges can best match experts? Is usefulness consistently difficult? This also raises the question of whether these results would hold when using different vision-language models, particularly open-source alternatives.

A further consideration is the reliability of AI-based evaluation methods when \textit{experts themselves} diverge on certain design aspects. Since experts may disagree on criteria such as creativity or functionality of a given design, how and how should an AI model predict expert ratings? Future work should examine inter-expert variability, perhaps using ensemble AI methods or probabilistic models to capture the distribution of expert ratings and provide uncertainty estimates in AI judgments.

There are important ethical considerations to be made around the use of AI as an evaluator. We aim to highlight the importance of ensuring that an AI's evaluations align with human judgment to mitigate risk, but other factors must also be considered. For instance, an AI judge could amplify expert biases at scale. Moreover, while an AI model may outperform a novice in replicating expert ratings, this does not diminish the value of novice training. It remains crucial to determine which skills engineers and designers should retain and develop, regardless of AI’s capabilities. Additionally, we must examine when and how large-scale design evaluation should occur. While this has been explored for decades—such as in Google's Project 10 to the 100th—AI has dramatically increased the volume of generated designs, raising new questions about its role in evaluating this growing abundance.

\section{Conclusion}

We proposed using \emph{in-context} vision-language models---which do not undergo additional training---to rate design sketches on subjective metrics like uniqueness, usefulness, and creativity. To thoroughly verify when such an AI truly performs at a ``human expert'' level, we combined multiple statistical tests: inter-rater agreement, error analysis, distributional checks, correlation, top-set overlap, \emph{and} equivalence (TOST). This ensures that claims of ``no difference'' are supported by \emph{actual evidence of equivalence} rather than the mere absence of significance. Our findings suggest that AI judges can serve as viable substitutes for trained novices, outperforming them in several cases and providing scalable, cost-efficient alternatives for large-scale evaluations. Furthermore, our experiments demonstrate that, with carefully constructed few-shot prompting, a reasoning-supported VLM \emph{is} expert-equivalent for design metrics such as uniqueness and drawing quality. We conclude that \emph{both} in-context learning and rigorous multi-metric, multi-test statistical validation are crucial to confidently deploy AI judges in design. Moving forward, we believe this holistic approach will generalize to other subjective domains, guiding practitioners to adopt AI safely and effectively as a peer to human experts.




\bibliographystyle{asmeconf}   

\bibliography{main} 


\appendix
\section{Additional Runs for All Metrics}

        \begin{table*}[!htbp]\centering
            \caption{Run 2: Summary of Uniqueness results. Kappa and ICC are relative to the expert--expert baseline of 0.53. MAE values compare to the observed Expert--Expert MAE = 1.11. ``Equiv?'' indicates whether TOST found AI--expert equivalence within $\pm1.0$ margin. Bold indicates judge is within 20\% of expert-expert agreement.}
            \label{tab:uniqueness_run2}
            \begin{tabular}{ll c ccc cccc c}
            \toprule
              & \textbf{Comparison} & \textbf{Kappa} & \textbf{ICC} & \textbf{MAE} & \textbf{\shortstack{Mean Diff \\ $\pm$ 1.96 SD}} & \textbf{Equiv?} & \textbf{Spear.} & \textbf{\shortstack{Wilcoxon \\ p-corr}} & \textbf{\shortstack{Jacc. \\ AUC}} & \textbf{\shortstack{Tests\\ Passed}} \\
            \midrule
            & \textbf{Expert 1 vs Expert 2} & 0.53 & 0.53 & 1.11 & $0.33 \pm 2.86$ & -- & 0.53 & $ 3.09 \cdot 10^{-09 }$ & 0.64 & -- \\
            \midrule
        \midrule\multirow{7}{*}{\rotatebox{90}{\textbf{Expert 1}}}
& Trained Novice 1 & \textbf{0.49} & \textbf{0.49} & \textbf{1.17} & $\textbf{-0.31} \pm \textbf{3.01}$ & \textbf{True} & \textbf{0.48} & $ 1.83 \cdot 10^{-08 }$ & \textbf{0.63} & 8/9 \\ 
& Trained Novice 2 & \textbf{0.44} & \textbf{0.44} & \textbf{1.21} & $0.83 \pm \textbf{2.80}$ & \textbf{True} & \textbf{0.49} & $ 2.32 \cdot 10^{-49 }$ & \textbf{0.63} & 7/9 \\ 
& Trained Novice 3 & \textbf{0.51} & \textbf{0.51} & \textbf{1.04} & $\textbf{-0.10} \pm \textbf{2.68}$ & \textbf{True} & \textbf{0.49} & $\mathbf{1.57 \cdot 10^{-01}}$ & \textbf{0.64} & 9/9 \\ 
& AI Judge: No Context & 0.22 & 0.21 & 1.61 & $-1.39 \pm \textbf{2.78}$ & False & 0.38 & $ 1.27 \cdot 10^{-94 }$ & \textbf{0.60} & 2/9 \\ 
& AI Judge: Text & 0.38 & 0.38 & \textbf{1.20} & $\textbf{0.13} \pm \textbf{3.10}$ & \textbf{True} & 0.37 & $\mathbf{7.67 \cdot 10^{-02}}$ & \textbf{0.59} & 6/9 \\ 
& AI Judge: Text + Image & \textbf{0.45} & \textbf{0.45} & \textbf{1.14} & $\textbf{-0.37} \pm \textbf{2.79}$ & \textbf{True} & \textbf{0.47} & $ 3.08 \cdot 10^{-12 }$ & \textbf{0.63} & 8/9 \\ 
& AI Judge: T + I + Reasoning & \textbf{0.52} & \textbf{0.52} & \textbf{0.99} & $\textbf{-0.16} \pm \textbf{2.57}$ & \textbf{True} & \textbf{0.53} & $ 1.45 \cdot 10^{-03 }$ & \textbf{0.64} & 8/9 \\ 
\midrule\multirow{7}{*}{\rotatebox{90}{\textbf{Expert 2}}}
& Trained Novice 1 & \textbf{0.58} & \textbf{0.58} & \textbf{1.07} & $\textbf{0.01} \pm \textbf{2.83}$ & \textbf{True} & \textbf{0.56} & $\mathbf{1.00 \cdot 10^{+00}}$ & \textbf{0.65} & 9/9 \\ 
& Trained Novice 2 & 0.42 & 0.42 & 1.40 & $1.15 \pm \textbf{2.78}$ & False & \textbf{0.53} & $ 8.69 \cdot 10^{-77 }$ & \textbf{0.66} & 3/9 \\ 
& Trained Novice 3 & \textbf{0.55} & \textbf{0.55} & \textbf{1.00} & $\textbf{0.23} \pm \textbf{2.62}$ & \textbf{True} & \textbf{0.55} & $ 1.11 \cdot 10^{-06 }$ & \textbf{0.65} & 8/9 \\ 
& AI Judge: No Context & 0.26 & 0.26 & 1.45 & $-1.06 \pm \textbf{2.87}$ & False & 0.41 & $ 1.52 \cdot 10^{-67 }$ & \textbf{0.61} & 2/9 \\ 
& AI Judge: Text & 0.41 & 0.41 & \textbf{1.19} & $0.46 \pm \textbf{3.06}$ & \textbf{True} & 0.42 & $ 7.28 \cdot 10^{-15 }$ & \textbf{0.63} & 4/9 \\ 
& AI Judge: Text + Image & \textbf{0.53} & \textbf{0.53} & \textbf{1.04} & $\textbf{-0.04} \pm \textbf{2.70}$ & \textbf{True} & \textbf{0.54} & $\mathbf{6.84 \cdot 10^{-01}}$ & \textbf{0.65} & 9/9 \\ 
& AI Judge: T + I + Reasoning & \textbf{0.54} & \textbf{0.54} & \textbf{0.99} & $\textbf{0.17} \pm \textbf{2.60}$ & \textbf{True} & \textbf{0.56} & $ 1.36 \cdot 10^{-03 }$ & \textbf{0.65} & 8/9 \\ 
\bottomrule
\end{tabular}
\end{table*}

        \begin{table*}[!ht]\centering
            \caption{Run 3: Summary of Uniqueness results. Kappa and ICC are relative to the expert--expert baseline of 0.53. MAE values compare to the observed Expert--Expert MAE = 1.11. ``Equiv?'' indicates whether TOST found AI--expert equivalence within $\pm1.0$ margin. Bold indicates judge is within 20\% of expert-expert agreement.}
            \label{tab:uniqueness_run3}
            \begin{tabular}{ll c ccc cccc c}
            \toprule
              & \textbf{Comparison} & \textbf{Kappa} & \textbf{ICC} & \textbf{MAE} & \textbf{\shortstack{Mean Diff \\ $\pm$ 1.96 SD}} & \textbf{Equiv?} & \textbf{Spear.} & \textbf{\shortstack{Wilcoxon \\ p-corr}} & \textbf{\shortstack{Jacc. \\ AUC}} & \textbf{\shortstack{Tests\\ Passed}} \\
            \midrule
            & \textbf{Expert 1 vs Expert 2} & 0.53 & 0.54 & 1.11 & $0.33 \pm 2.86$ & -- & 0.53 & $ 3.16 \cdot 10^{-09 }$ & 0.64 & -- \\
            \midrule
        \midrule\multirow{7}{*}{\rotatebox{90}{\textbf{Expert 1}}}
& Trained Novice 1 & \textbf{0.49} & \textbf{0.49} & \textbf{1.17} & $\textbf{-0.32} \pm \textbf{3.01}$ & \textbf{True} & \textbf{0.48} & $ 1.36 \cdot 10^{-08 }$ & \textbf{0.63} & 8/9 \\ 
& Trained Novice 2 & \textbf{0.44} & \textbf{0.44} & \textbf{1.20} & $0.83 \pm \textbf{2.80}$ & \textbf{True} & \textbf{0.49} & $ 1.78 \cdot 10^{-49 }$ & \textbf{0.63} & 7/9 \\ 
& Trained Novice 3 & \textbf{0.51} & \textbf{0.51} & \textbf{1.05} & $\textbf{-0.11} \pm \textbf{2.67}$ & \textbf{True} & \textbf{0.50} & $\mathbf{9.43 \cdot 10^{-02}}$ & \textbf{0.64} & 9/9 \\ 
& AI Judge: No Context & 0.23 & 0.23 & 1.61 & $-1.36 \pm \textbf{2.76}$ & False & 0.41 & $ 3.37 \cdot 10^{-92 }$ & \textbf{0.61} & 2/9 \\ 
& AI Judge: Text & 0.41 & 0.41 & \textbf{1.27} & $\textbf{-0.16} \pm \textbf{3.29}$ & \textbf{True} & 0.41 & $ 1.70 \cdot 10^{-02 }$ & \textbf{0.61} & 5/9 \\ 
& AI Judge: Text + Image & \textbf{0.45} & \textbf{0.46} & \textbf{1.23} & $-0.57 \pm \textbf{2.93}$ & \textbf{True} & \textbf{0.49} & $ 1.82 \cdot 10^{-25 }$ & \textbf{0.62} & 7/9 \\ 
& AI Judge: T + I + Reasoning & \textbf{0.52} & \textbf{0.52} & \textbf{1.06} & $\textbf{-0.36} \pm \textbf{2.67}$ & \textbf{True} & \textbf{0.53} & $ 6.66 \cdot 10^{-13 }$ & \textbf{0.64} & 8/9 \\ 
\midrule\multirow{7}{*}{\rotatebox{90}{\textbf{Expert 2}}}
& Trained Novice 1 & \textbf{0.58} & \textbf{0.58} & \textbf{1.07} & $\textbf{0.01} \pm \textbf{2.83}$ & \textbf{True} & \textbf{0.56} & $\mathbf{1.00 \cdot 10^{+00}}$ & \textbf{0.65} & 9/9 \\ 
& Trained Novice 2 & 0.42 & 0.42 & 1.40 & $1.15 \pm \textbf{2.77}$ & False & \textbf{0.52} & $ 8.57 \cdot 10^{-77 }$ & \textbf{0.66} & 3/9 \\ 
& Trained Novice 3 & \textbf{0.54} & \textbf{0.54} & \textbf{1.00} & $\textbf{0.22} \pm \textbf{2.63}$ & \textbf{True} & \textbf{0.55} & $ 3.25 \cdot 10^{-06 }$ & \textbf{0.65} & 8/9 \\ 
& AI Judge: No Context & 0.26 & 0.26 & 1.45 & $-1.03 \pm \textbf{2.89}$ & False & \textbf{0.42} & $ 8.40 \cdot 10^{-66 }$ & \textbf{0.62} & 3/9 \\ 
& AI Judge: Text & \textbf{0.45} & \textbf{0.45} & \textbf{1.24} & $\textbf{0.17} \pm \textbf{3.25}$ & \textbf{True} & \textbf{0.45} & $ 4.60 \cdot 10^{-02 }$ & \textbf{0.62} & 8/9 \\ 
& AI Judge: Text + Image & \textbf{0.52} & \textbf{0.52} & \textbf{1.13} & $\textbf{-0.24} \pm \textbf{2.88}$ & \textbf{True} & \textbf{0.54} & $ 1.68 \cdot 10^{-07 }$ & \textbf{0.63} & 8/9 \\ 
& AI Judge: T + I + Reasoning & \textbf{0.55} & \textbf{0.55} & \textbf{1.01} & $\textbf{-0.03} \pm \textbf{2.70}$ & \textbf{True} & \textbf{0.56} & $\mathbf{9.71 \cdot 10^{-01}}$ & \textbf{0.65} & 9/9 \\ 
\bottomrule
\end{tabular}
\end{table*}

        \begin{table*}[!ht]\centering
            \caption{Run 2: Summary of Creativity results. Kappa and ICC are relative to the expert--expert baseline of 0.26. MAE values compare to the observed Expert--Expert MAE = 1.25. ``Equiv?'' indicates whether TOST found AI--expert equivalence within $\pm1.0$ margin. Bold indicates judge is within 20\% of expert-expert agreement.}
            \label{tab:creativity_run2}
            \begin{tabular}{ll c ccc cccc c}
            \toprule
              & \textbf{Comparison} & \textbf{Kappa} & \textbf{ICC} & \textbf{MAE} & \textbf{\shortstack{Mean Diff \\ $\pm$ 1.96 SD}} & \textbf{Equiv?} & \textbf{Spear.} & \textbf{\shortstack{Wilcoxon \\ p-corr}} & \textbf{\shortstack{Jacc. \\ AUC}} & \textbf{\shortstack{Tests\\ Passed}} \\
            \midrule
            & \textbf{Expert 1 vs Expert 2} & 0.26 & 0.26 & 1.25 & $0.19 \pm 3.13$ & -- & 0.22 & $ 1.98 \cdot 10^{-03 }$ & 0.58 & -- \\
            \midrule
        \midrule\multirow{7}{*}{\rotatebox{90}{\textbf{Expert 1}}}
& Trained Novice 1 & -0.02 & -0.02 & 1.62 & $-0.89 \pm \textbf{3.59}$ & \textbf{True} & -0.04 & $ 2.83 \cdot 10^{-37 }$ & \textbf{0.55} & 3/9 \\ 
& Trained Novice 2 & 0.09 & 0.09 & \textbf{1.38} & $0.59 \pm \textbf{3.39}$ & \textbf{True} & 0.12 & $ 8.63 \cdot 10^{-22 }$ & \textbf{0.57} & 4/9 \\ 
& Trained Novice 3 & -0.07 & -0.07 & 1.51 & $-0.58 \pm \textbf{3.64}$ & \textbf{True} & -0.07 & $ 5.81 \cdot 10^{-17 }$ & \textbf{0.57} & 3/9 \\ 
& AI Judge: No Context & 0.05 & 0.05 & 2.03 & $-1.90 \pm \textbf{2.93}$ & False & 0.11 & $ 5.21 \cdot 10^{-112 }$ & \textbf{0.58} & 2/9 \\ 
& AI Judge: Text & 0.09 & 0.09 & \textbf{1.46} & $-0.66 \pm \textbf{3.36}$ & \textbf{True} & 0.09 & $ 4.06 \cdot 10^{-25 }$ & \textbf{0.56} & 4/9 \\ 
& AI Judge: Text + Image & 0.11 & 0.11 & 1.51 & $-1.09 \pm \textbf{3.03}$ & False & 0.14 & $ 1.02 \cdot 10^{-63 }$ & \textbf{0.59} & 2/9 \\ 
& AI Judge: T + I + Reasoning & 0.11 & 0.11 & \textbf{1.40} & $-0.95 \pm \textbf{2.98}$ & False & 0.14 & $ 6.10 \cdot 10^{-54 }$ & \textbf{0.57} & 3/9 \\ 
\midrule\multirow{7}{*}{\rotatebox{90}{\textbf{Expert 2}}}
& Trained Novice 1 & 0.17 & 0.17 & \textbf{1.32} & $-0.70 \pm \textbf{3.03}$ & \textbf{True} & \textbf{0.19} & $ 6.63 \cdot 10^{-33 }$ & \textbf{0.60} & 5/9 \\ 
& Trained Novice 2 & 0.16 & 0.16 & \textbf{1.34} & $0.78 \pm \textbf{3.08}$ & \textbf{True} & \textbf{0.18} & $ 3.49 \cdot 10^{-39 }$ & \textbf{0.60} & 5/9 \\ 
& Trained Novice 3 & \textbf{0.23} & \textbf{0.23} & \textbf{1.20} & $-0.39 \pm \textbf{2.92}$ & \textbf{True} & \textbf{0.25} & $ 2.60 \cdot 10^{-12 }$ & \textbf{0.59} & 7/9 \\ 
& AI Judge: No Context & 0.08 & 0.08 & 1.84 & $-1.71 \pm \textbf{2.66}$ & False & \textbf{0.19} & $ 2.55 \cdot 10^{-112 }$ & \textbf{0.57} & 3/9 \\ 
& AI Judge: Text & 0.14 & 0.14 & \textbf{1.29} & $-0.47 \pm \textbf{3.12}$ & \textbf{True} & 0.16 & $ 6.09 \cdot 10^{-16 }$ & \textbf{0.58} & 4/9 \\ 
& AI Judge: Text + Image & 0.17 & 0.17 & \textbf{1.33} & $-0.90 \pm \textbf{2.75}$ & \textbf{True} & \textbf{0.22} & $ 1.11 \cdot 10^{-55 }$ & \textbf{0.57} & 5/9 \\ 
& AI Judge: T + I + Reasoning & \textbf{0.24} & \textbf{0.24} & \textbf{1.17} & $-0.75 \pm \textbf{2.56}$ & \textbf{True} & \textbf{0.31} & $ 3.93 \cdot 10^{-48 }$ & \textbf{0.60} & 7/9 \\ 
\bottomrule
\end{tabular}
\end{table*}

        \begin{table*}[!ht]\centering
            \caption{Run 3: Summary of Creativity results. Kappa and ICC are relative to the expert--expert baseline of 0.26. MAE values compare to the observed Expert--Expert MAE = 1.25. ``Equiv?'' indicates whether TOST found AI--expert equivalence within $\pm1.0$ margin. Bold indicates judge is within 20\% of expert-expert agreement.}
            \label{tab:creativity_run3}
            \begin{tabular}{ll c ccc cccc c}
            \toprule
              & \textbf{Comparison} & \textbf{Kappa} & \textbf{ICC} & \textbf{MAE} & \textbf{\shortstack{Mean Diff \\ $\pm$ 1.96 SD}} & \textbf{Equiv?} & \textbf{Spear.} & \textbf{\shortstack{Wilcoxon \\ p-corr}} & \textbf{\shortstack{Jacc. \\ AUC}} & \textbf{\shortstack{Tests\\ Passed}} \\
            \midrule
            & \textbf{Expert 1 vs Expert 2} & 0.26 & 0.26 & 1.25 & $0.18 \pm 3.14$ & -- & 0.22 & $ 3.83 \cdot 10^{-03 }$ & 0.58 & -- \\
            \midrule
        \midrule\multirow{7}{*}{\rotatebox{90}{\textbf{Expert 1}}}
& Trained Novice 1 & -0.03 & -0.03 & 1.63 & $-0.88 \pm \textbf{3.61}$ & \textbf{True} & -0.05 & $ 2.23 \cdot 10^{-36 }$ & \textbf{0.55} & 3/9 \\ 
& Trained Novice 2 & 0.08 & 0.09 & \textbf{1.38} & $0.60 \pm \textbf{3.39}$ & \textbf{True} & 0.11 & $ 1.61 \cdot 10^{-22 }$ & \textbf{0.57} & 4/9 \\ 
& Trained Novice 3 & -0.06 & -0.06 & \textbf{1.50} & $-0.58 \pm \textbf{3.62}$ & \textbf{True} & -0.07 & $ 4.43 \cdot 10^{-17 }$ & \textbf{0.57} & 4/9 \\ 
& AI Judge: No Context & 0.06 & 0.06 & 2.01 & $-1.89 \pm \textbf{2.92}$ & False & 0.11 & $ 2.00 \cdot 10^{-113 }$ & \textbf{0.58} & 2/9 \\ 
& AI Judge: Text & 0.11 & 0.11 & \textbf{1.46} & $-0.87 \pm \textbf{3.19}$ & \textbf{True} & 0.13 & $ 6.16 \cdot 10^{-43 }$ & \textbf{0.56} & 4/9 \\ 
& AI Judge: Text + Image & 0.10 & 0.10 & 1.51 & $-1.09 \pm \textbf{3.02}$ & False & 0.14 & $ 1.34 \cdot 10^{-64 }$ & \textbf{0.59} & 2/9 \\ 
& AI Judge: T + I + Reasoning & 0.12 & 0.12 & \textbf{1.47} & $-1.10 \pm \textbf{2.88}$ & False & \textbf{0.19} & $ 1.21 \cdot 10^{-69 }$ & \textbf{0.60} & 4/9 \\ 
\midrule\multirow{7}{*}{\rotatebox{90}{\textbf{Expert 2}}}
& Trained Novice 1 & 0.17 & 0.17 & \textbf{1.32} & $-0.70 \pm \textbf{3.04}$ & \textbf{True} & \textbf{0.19} & $ 8.86 \cdot 10^{-33 }$ & \textbf{0.60} & 5/9 \\ 
& Trained Novice 2 & 0.15 & 0.15 & \textbf{1.33} & $0.78 \pm \textbf{3.07}$ & \textbf{True} & \textbf{0.18} & $ 1.18 \cdot 10^{-39 }$ & \textbf{0.60} & 5/9 \\ 
& Trained Novice 3 & \textbf{0.23} & \textbf{0.23} & \textbf{1.20} & $-0.39 \pm \textbf{2.91}$ & \textbf{True} & \textbf{0.25} & $ 7.72 \cdot 10^{-13 }$ & \textbf{0.59} & 7/9 \\ 
& AI Judge: No Context & 0.10 & 0.10 & 1.82 & $-1.71 \pm \textbf{2.62}$ & False & \textbf{0.22} & $ 8.94 \cdot 10^{-115 }$ & \textbf{0.58} & 3/9 \\ 
& AI Judge: Text & 0.14 & 0.14 & \textbf{1.33} & $-0.69 \pm \textbf{3.00}$ & \textbf{True} & 0.15 & $ 3.42 \cdot 10^{-32 }$ & \textbf{0.59} & 4/9 \\ 
& AI Judge: Text + Image & 0.13 & 0.13 & \textbf{1.36} & $-0.91 \pm \textbf{2.82}$ & \textbf{True} & 0.16 & $ 6.29 \cdot 10^{-55 }$ & \textbf{0.57} & 4/9 \\ 
& AI Judge: T + I + Reasoning & 0.16 & 0.16 & \textbf{1.31} & $-0.91 \pm \textbf{2.66}$ & \textbf{True} & \textbf{0.22} & $ 1.19 \cdot 10^{-59 }$ & \textbf{0.58} & 5/9 \\ 
\bottomrule
\end{tabular}
\end{table*}

        \begin{table*}[!ht]\centering
            \caption{Run 2: Summary of Usefulness results. Kappa and ICC are relative to the expert--expert baseline of 0.58. MAE values compare to the observed Expert--Expert MAE = 1.00. ``Equiv?'' indicates whether TOST found AI--expert equivalence within $\pm1.0$ margin. Bold indicates judge is within 20\% of expert-expert agreement.}
            \label{tab:usefulness_run2}
            \begin{tabular}{ll c ccc cccc c}
            \toprule
              & \textbf{Comparison} & \textbf{Kappa} & \textbf{ICC} & \textbf{MAE} & \textbf{\shortstack{Mean Diff \\ $\pm$ 1.96 SD}} & \textbf{Equiv?} & \textbf{Spear.} & \textbf{\shortstack{Wilcoxon \\ p-corr}} & \textbf{\shortstack{Jacc. \\ AUC}} & \textbf{\shortstack{Tests\\ Passed}} \\
            \midrule
            & \textbf{Expert 1 vs Expert 2} & 0.58 & 0.60 & 1.00 & $-0.01 \pm 2.61$ & -- & 0.58 & $\mathbf{1.00 \cdot 10^{+00}}$ & 0.67 & -- \\
            \midrule
        \midrule\multirow{7}{*}{\rotatebox{90}{\textbf{Expert 1}}}
& Trained Novice 1 & 0.11 & 0.11 & 1.53 & $-0.81 \pm 1.68$ & \textbf{True} & 0.14 & $ 3.60 \cdot 10^{-35 }$ & \textbf{0.55} & 3/9 \\ 
& Trained Novice 2 & 0.35 & 0.35 & 1.60 & $-1.21 \pm 1.65$ & False & 0.45 & $ 3.53 \cdot 10^{-67 }$ & \textbf{0.62} & 2/9 \\ 
& Trained Novice 3 & 0.35 & 0.35 & 1.21 & $-0.36 \pm \textbf{2.93}$ & \textbf{True} & 0.36 & $ 1.47 \cdot 10^{-11 }$ & \textbf{0.60} & 4/9 \\ 
& AI Judge: No Context & 0.29 & 0.28 & 1.35 & $-0.82 \pm \textbf{2.77}$ & \textbf{True} & 0.36 & $ 7.98 \cdot 10^{-48 }$ & \textbf{0.60} & 4/9 \\ 
& AI Judge: Text & 0.27 & 0.27 & 1.28 & $-0.16 \pm 1.61$ & \textbf{True} & 0.28 & $ 4.13 \cdot 10^{-03 }$ & \textbf{0.57} & 3/9 \\ 
& AI Judge: Text + Image & 0.25 & 0.24 & 1.32 & $-0.32 \pm 1.61$ & \textbf{True} & 0.25 & $ 6.12 \cdot 10^{-08 }$ & \textbf{0.57} & 3/9 \\ 
& AI Judge: T + I + Reasoning & 0.30 & 0.30 & 1.31 & $-0.71 \pm \textbf{2.84}$ & \textbf{True} & 0.34 & $ 1.05 \cdot 10^{-36 }$ & \textbf{0.59} & 4/9 \\ 
\midrule\multirow{7}{*}{\rotatebox{90}{\textbf{Expert 2}}}
& Trained Novice 1 & 0.15 & 0.14 & 1.54 & $-0.82 \pm 1.69$ & \textbf{True} & 0.17 & $ 1.48 \cdot 10^{-36 }$ & \textbf{0.54} & 3/9 \\ 
& Trained Novice 2 & 0.33 & 0.33 & 1.70 & $-1.22 \pm 1.72$ & False & 0.43 & $ 2.11 \cdot 10^{-63 }$ & \textbf{0.59} & 2/9 \\ 
& Trained Novice 3 & 0.24 & 0.25 & 1.38 & $-0.37 \pm 1.66$ & \textbf{True} & 0.22 & $ 1.33 \cdot 10^{-09 }$ & \textbf{0.56} & 3/9 \\ 
& AI Judge: No Context & 0.16 & 0.16 & 1.48 & $-0.83 \pm 1.62$ & \textbf{True} & 0.18 & $ 3.97 \cdot 10^{-40 }$ & \textbf{0.55} & 3/9 \\ 
& AI Judge: Text & 0.12 & 0.12 & 1.48 & $-0.17 \pm 1.81$ & \textbf{True} & 0.11 & $ 2.28 \cdot 10^{-02 }$ & 0.527 & 2/9 \\ 
& AI Judge: Text + Image & 0.08 & 0.09 & 1.53 & $-0.33 \pm 1.82$ & \textbf{True} & 0.06 & $ 6.16 \cdot 10^{-07 }$ & 0.528 & 2/9 \\ 
& AI Judge: T + I + Reasoning & 0.16 & 0.16 & 1.48 & $-0.72 \pm 1.66$ & \textbf{True} & 0.15 & $ 8.58 \cdot 10^{-31 }$ & \textbf{0.54} & 3/9 \\ 
\bottomrule
\end{tabular}
\end{table*}

        \begin{table*}[!ht]\centering
            \caption{Run 3: Summary of Usefulness results. Kappa and ICC are relative to the expert--expert baseline of 0.58. MAE values compare to the observed Expert--Expert MAE = 1.00. ``Equiv?'' indicates whether TOST found AI--expert equivalence within $\pm1.0$ margin. Bold indicates judge is within 20\% of expert-expert agreement.}
            \label{tab:usefulness_run3}
            \begin{tabular}{ll c ccc cccc c}
            \toprule
              & \textbf{Comparison} & \textbf{Kappa} & \textbf{ICC} & \textbf{MAE} & \textbf{\shortstack{Mean Diff \\ $\pm$ 1.96 SD}} & \textbf{Equiv?} & \textbf{Spear.} & \textbf{\shortstack{Wilcoxon \\ p-corr}} & \textbf{\shortstack{Jacc. \\ AUC}} & \textbf{\shortstack{Tests\\ Passed}} \\
            \midrule
            & \textbf{Expert 1 vs Expert 2} & 0.58 & 0.59 & 1.00 & $-0.01 \pm 2.61$ & -- & 0.58 & $\mathbf{1.00 \cdot 10^{+00}}$ & 0.67 & -- \\
            \midrule
        \midrule\multirow{7}{*}{\rotatebox{90}{\textbf{Expert 1}}}
& Trained Novice 1 & 0.12 & 0.12 & 1.53 & $-0.82 \pm 1.67$ & \textbf{True} & 0.15 & $ 4.05 \cdot 10^{-36 }$ & \textbf{0.56} & 3/9 \\ 
& Trained Novice 2 & 0.35 & 0.35 & 1.60 & $-1.20 \pm 1.65$ & False & 0.45 & $ 6.06 \cdot 10^{-67 }$ & \textbf{0.63} & 2/9 \\ 
& Trained Novice 3 & 0.35 & 0.36 & \textbf{1.19} & $-0.36 \pm \textbf{2.92}$ & \textbf{True} & 0.36 & $ 1.10 \cdot 10^{-11 }$ & \textbf{0.62} & 5/9 \\ 
& AI Judge: No Context & 0.28 & 0.28 & 1.36 & $-0.82 \pm \textbf{2.77}$ & \textbf{True} & 0.36 & $ 4.25 \cdot 10^{-49 }$ & \textbf{0.61} & 4/9 \\ 
& AI Judge: Text & 0.32 & 0.32 & 1.32 & $-0.62 \pm \textbf{2.98}$ & \textbf{True} & 0.35 & $ 6.73 \cdot 10^{-29 }$ & \textbf{0.61} & 4/9 \\ 
& AI Judge: Text + Image & 0.32 & 0.32 & 1.35 & $-0.64 \pm \textbf{3.05}$ & \textbf{True} & 0.35 & $ 4.06 \cdot 10^{-28 }$ & \textbf{0.59} & 4/9 \\ 
& AI Judge: T + I + Reasoning & 0.32 & 0.31 & 1.40 & $-0.99 \pm \textbf{2.73}$ & False & 0.40 & $ 8.73 \cdot 10^{-65 }$ & \textbf{0.62} & 3/9 \\ 
\midrule\multirow{7}{*}{\rotatebox{90}{\textbf{Expert 2}}}
& Trained Novice 1 & 0.15 & 0.15 & 1.55 & $-0.83 \pm 1.70$ & \textbf{True} & 0.17 & $ 2.69 \cdot 10^{-37 }$ & \textbf{0.54} & 3/9 \\ 
& Trained Novice 2 & 0.33 & 0.33 & 1.71 & $-1.22 \pm 1.73$ & False & 0.43 & $ 4.15 \cdot 10^{-63 }$ & \textbf{0.59} & 2/9 \\ 
& Trained Novice 3 & 0.24 & 0.25 & 1.38 & $-0.37 \pm 1.67$ & \textbf{True} & 0.22 & $ 1.22 \cdot 10^{-09 }$ & \textbf{0.56} & 3/9 \\ 
& AI Judge: No Context & 0.14 & 0.15 & 1.49 & $-0.84 \pm 1.63$ & \textbf{True} & 0.17 & $ 1.40 \cdot 10^{-40 }$ & \textbf{0.55} & 3/9 \\ 
& AI Judge: Text & 0.15 & 0.16 & 1.53 & $-0.64 \pm 1.76$ & \textbf{True} & 0.16 & $ 1.21 \cdot 10^{-21 }$ & \textbf{0.54} & 3/9 \\ 
& AI Judge: Text + Image & 0.16 & 0.16 & 1.56 & $-0.66 \pm 1.80$ & \textbf{True} & 0.16 & $ 1.08 \cdot 10^{-22 }$ & \textbf{0.55} & 3/9 \\ 
& AI Judge: T + I + Reasoning & 0.17 & 0.17 & 1.57 & $-1.01 \pm 1.64$ & False & 0.21 & $ 3.38 \cdot 10^{-52 }$ & \textbf{0.55} & 2/9 \\ 
\bottomrule
\end{tabular}
\end{table*}

       \begin{table*}[!ht]\centering
            \caption{Run 2: Summary of Drawing Quality results. Kappa and ICC are relative to the expert--expert baseline of 0.33. MAE values compare to the observed Expert--Expert MAE = 1.16. ``Equiv?'' indicates whether TOST found AI--expert equivalence within $\pm1.0$ margin. Bold indicates judge is within 20\% of expert-expert agreement.}
            \label{tab:drawing_run2}
            \begin{tabular}{ll c ccc cccc c}
            \toprule
              & \textbf{Comparison} & \textbf{Kappa} & \textbf{ICC} & \textbf{MAE} & \textbf{\shortstack{Mean Diff \\ $\pm$ 1.96 SD}} & \textbf{Equiv?} & \textbf{Spear.} & \textbf{\shortstack{Wilcoxon \\ p-corr}} & \textbf{\shortstack{Jacc. \\ AUC}} & \textbf{\shortstack{Tests\\ Passed}} \\
            \midrule
            & \textbf{Expert 1 vs Expert 2} & 0.33 & 0.33 & 1.16 & $-0.67 \pm 2.64$ & -- & 0.37 & $ 8.09 \cdot 10^{-39 }$ & 0.61 & -- \\
            \midrule
        \midrule\multirow{7}{*}{\rotatebox{90}{\textbf{Expert 1}}}
& Trained Novice 1 & 0.16 & 0.16 & \textbf{1.26} & $\textbf{0.45} \pm \textbf{2.99}$ & \textbf{True} & 0.17 & $ 7.45 \cdot 10^{-16 }$ & \textbf{0.58} & 5/9 \\ 
& Trained Novice 2 & 0.23 & 0.23 & 1.46 & $0.93 \pm \textbf{3.08}$ & False & 0.27 & $ 1.37 \cdot 10^{-48 }$ & \textbf{0.60} & 2/9 \\ 
& Trained Novice 3 & 0.21 & 0.22 & \textbf{1.23} & $0.82 \pm \textbf{2.63}$ & \textbf{True} & 0.25 & $ 1.95 \cdot 10^{-52 }$ & \textbf{0.58} & 4/9 \\ 
& AI Judge: No Context & \textbf{0.28} & \textbf{0.29} & \textbf{1.04} & $\textbf{0.52} \pm \textbf{2.36}$ & \textbf{True} & \textbf{0.34} & $ 2.33 \cdot 10^{-31 }$ & \textbf{0.62} & 8/9 \\ 
& AI Judge: Text & 0.18 & 0.18 & \textbf{1.19} & $\textbf{0.58} \pm \textbf{2.83}$ & \textbf{True} & 0.20 & $ 3.68 \cdot 10^{-27 }$ & \textbf{0.58} & 5/9 \\ 
& AI Judge: Text + Image & 0.22 & 0.22 & \textbf{1.11} & $\textbf{0.48} \pm \textbf{2.68}$ & \textbf{True} & 0.25 & $ 2.69 \cdot 10^{-21 }$ & \textbf{0.60} & 5/9 \\ 
& AI Judge: T + I + Reasoning & \textbf{0.30} & \textbf{0.30} & \textbf{0.96} & $\textbf{0.28} \pm \textbf{2.38}$ & \textbf{True} & \textbf{0.33} & $ 1.55 \cdot 10^{-10 }$ & \textbf{0.62} & 8/9 \\ 
\midrule\multirow{7}{*}{\rotatebox{90}{\textbf{Expert 2}}}
& Trained Novice 1 & \textbf{0.36} & \textbf{0.36} & \textbf{0.98} & $\textbf{-0.23} \pm \textbf{2.54}$ & \textbf{True} & \textbf{0.35} & $ 5.31 \cdot 10^{-06 }$ & \textbf{0.62} & 8/9 \\ 
& Trained Novice 2 & \textbf{0.44} & \textbf{0.44} & \textbf{1.02} & $\textbf{0.25} \pm \textbf{2.64}$ & \textbf{True} & \textbf{0.46} & $ 9.77 \cdot 10^{-08 }$ & \textbf{0.66} & 8/9 \\ 
& Trained Novice 3 & \textbf{0.43} & \textbf{0.43} & \textbf{0.87} & $\textbf{0.15} \pm \textbf{2.26}$ & \textbf{True} & \textbf{0.42} & $ 8.42 \cdot 10^{-04 }$ & \textbf{0.63} & 8/9 \\ 
& AI Judge: No Context & \textbf{0.41} & \textbf{0.41} & \textbf{0.81} & $\textbf{-0.15} \pm \textbf{2.12}$ & \textbf{True} & \textbf{0.45} & $ 2.44 \cdot 10^{-04 }$ & \textbf{0.65} & 8/9 \\ 
& AI Judge: Text & 0.09 & 0.09 & \textbf{1.15} & $\textbf{-0.09} \pm \textbf{2.93}$ & \textbf{True} & 0.10 & $\mathbf{1.25 \cdot 10^{-01}}$ & \textbf{0.56} & 6/9 \\ 
& AI Judge: Text + Image & 0.20 & 0.21 & \textbf{1.03} & $\textbf{-0.19} \pm \textbf{2.64}$ & \textbf{True} & 0.22 & $ 3.55 \cdot 10^{-05 }$ & \textbf{0.60} & 5/9 \\ 
& AI Judge: T + I + Reasoning & \textbf{0.29} & \textbf{0.29} & \textbf{0.93} & $\textbf{-0.39} \pm \textbf{2.28}$ & \textbf{True} & \textbf{0.34} & $ 4.09 \cdot 10^{-21 }$ & \textbf{0.64} & 8/9 \\ 
\bottomrule
\end{tabular}
\end{table*}

        \begin{table*}[!ht]\centering
            \caption{Run 3: Summary of Drawing Quality results. Kappa and ICC are relative to the expert--expert baseline of 0.33. MAE values compare to the observed Expert--Expert MAE = 1.16. ``Equiv?'' indicates whether TOST found AI--expert equivalence within $\pm1.0$ margin. Bold indicates judge is within 20\% of expert-expert agreement.}
            \label{tab:drawing_run3}
            \begin{tabular}{ll c ccc cccc c}
            \toprule
              & \textbf{Comparison} & \textbf{Kappa} & \textbf{ICC} & \textbf{MAE} & \textbf{\shortstack{Mean Diff \\ $\pm$ 1.96 SD}} & \textbf{Equiv?} & \textbf{Spear.} & \textbf{\shortstack{Wilcoxon \\ p-corr}} & \textbf{\shortstack{Jacc. \\ AUC}} & \textbf{\shortstack{Tests\\ Passed}} \\
            \midrule
            & \textbf{Expert 1 vs Expert 2} & 0.33 & 0.33 & 1.16 & $-0.67 \pm 2.66$ & -- & 0.37 & $ 1.65 \cdot 10^{-38 }$ & 0.62 & -- \\
            \midrule
        \midrule\multirow{7}{*}{\rotatebox{90}{\textbf{Expert 1}}}
& Trained Novice 1 & 0.17 & 0.17 & \textbf{1.26} & $\textbf{0.45} \pm \textbf{2.99}$ & \textbf{True} & 0.17 & $ 9.41 \cdot 10^{-16 }$ & \textbf{0.58} & 5/9 \\ 
& Trained Novice 2 & 0.23 & 0.23 & 1.46 & $0.92 \pm \textbf{3.08}$ & False & 0.27 & $ 4.82 \cdot 10^{-48 }$ & \textbf{0.61} & 2/9 \\ 
& Trained Novice 3 & 0.22 & 0.22 & \textbf{1.23} & $0.81 \pm \textbf{2.64}$ & \textbf{True} & 0.25 & $ 1.90 \cdot 10^{-51 }$ & \textbf{0.59} & 4/9 \\ 
& AI Judge: No Context & \textbf{0.27} & \textbf{0.27} & \textbf{1.04} & $\textbf{0.52} \pm \textbf{2.40}$ & \textbf{True} & \textbf{0.33} & $ 1.33 \cdot 10^{-29 }$ & \textbf{0.62} & 8/9 \\ 
& AI Judge: Text & 0.19 & 0.19 & \textbf{1.13} & $\textbf{0.00} \pm \textbf{2.87}$ & \textbf{True} & 0.19 & $\mathbf{1.00 \cdot 10^{+00}}$ & \textbf{0.56} & 6/9 \\ 
& AI Judge: Text + Image & \textbf{0.29} & \textbf{0.29} & \textbf{1.04} & $\textbf{0.01} \pm \textbf{2.62}$ & \textbf{True} & 0.28 & $\mathbf{1.00 \cdot 10^{+00}}$ & \textbf{0.61} & 8/9 \\ 
& AI Judge: T + I + Reasoning & \textbf{0.30} & \textbf{0.30} & \textbf{0.97} & $\textbf{-0.11} \pm \textbf{2.43}$ & \textbf{True} & \textbf{0.31} & $\mathbf{5.56 \cdot 10^{-02}}$ & \textbf{0.60} & 9/9 \\ 
\midrule\multirow{7}{*}{\rotatebox{90}{\textbf{Expert 2}}}

& Trained Novice 1 & \textbf{0.37} & \textbf{0.37} & \textbf{0.98} & $\textbf{-0.23} \pm \textbf{2.53}$ & \textbf{True} & \textbf{0.36} & $ 3.64 \cdot 10^{-06 }$ & \textbf{0.63} & 8/9 \\ 
& Trained Novice 2 & \textbf{0.45} & \textbf{0.45} & \textbf{1.02} & $\textbf{0.25} \pm \textbf{2.64}$ & \textbf{True} & \textbf{0.46} & $ 2.45 \cdot 10^{-07 }$ & \textbf{0.66} & 8/9 \\ 
& Trained Novice 3 & \textbf{0.43} & \textbf{0.43} & \textbf{0.87} & $\textbf{0.14} \pm \textbf{2.27}$ & \textbf{True} & \textbf{0.42} & $ 2.32 \cdot 10^{-03 }$ & \textbf{0.63} & 8/9 \\ 
& AI Judge: No Context & \textbf{0.37} & \textbf{0.37} & \textbf{0.84} & $\textbf{-0.16} \pm \textbf{2.19}$ & \textbf{True} & \textbf{0.42} & $ 2.46 \cdot 10^{-04 }$ & \textbf{0.65} & 8/9 \\ 
& AI Judge: Text & 0.07 & 0.07 & \textbf{1.31} & $\textbf{-0.67} \pm \textbf{2.97}$ & \textbf{True} & 0.09 & $ 6.95 \cdot 10^{-33 }$ & \textbf{0.60} & 5/9 \\ 
& AI Judge: Text + Image & 0.22 & 0.23 & \textbf{1.12} & $\textbf{-0.66} \pm \textbf{2.58}$ & \textbf{True} & 0.26 & $ 3.01 \cdot 10^{-40 }$ & \textbf{0.63} & 5/9 \\ 
& AI Judge: T + I + Reasoning & 0.24 & 0.24 & \textbf{1.11} & $\textbf{-0.78} \pm \textbf{2.32}$ & \textbf{True} & \textbf{0.35} & $ 1.27 \cdot 10^{-57 }$ & \textbf{0.61} & 6/9 \\ 
\bottomrule
\end{tabular}
\end{table*}

\end{document}